\documentclass[runningheads]{llncs}
\usepackage{graphicx}
\usepackage{amsmath,amssymb} 
\usepackage{color}
\usepackage[width=122mm,left=12mm,paperwidth=146mm,height=193mm,top=12mm,paperheight=217mm]{geometry}
\usepackage[titletoc]{appendix}
\usepackage{times}
\usepackage{epsfig}
\usepackage{graphicx}
\usepackage{amsmath}
\usepackage{amssymb}
\usepackage{siunitx} 
\usepackage{amsmath} 
\usepackage{comment}
\usepackage{amsmath}
\usepackage{caption}
\usepackage{subcaption}
\usepackage{color}
\usepackage{lipsum}
\usepackage{enumitem}
\usepackage{xr}
\usepackage{colortbl}
\usepackage{float}
\usepackage{authblk}
\usepackage{hyperref}

\newcommand{\eat}[1]{\ignorespaces}

\newcommand{\smallpar}[1]{\vspace{-6pt}\paragraph{\textbf{#1}}}

\begin{document}


\pagestyle{headings}
\mainmatter

\title{Learning to Globally Edit Images\\
with Textual Description} 



\author{Hai Wang {\footnote{
TTIC, Chicago, IL, 60637. USA. Email: haiwang@ttic.edu, work done at MSR }}
Jason D. Williams {\footnote{
Apple, Cupertino, CA, 95014. USA, Email: jdw@alumni.princeton.edu, work done at MSR}}
Sing Bing Kang {\footnote{
Microsoft Research, Redmond, WA, 98052, USA. Email: sbkang@microsoft.com}}
}

\maketitle

\begin{abstract}
	
	We show how we can globally edit images using textual instructions: given a source image and a textual instruction for the edit, generate a new image transformed under this instruction. To tackle this novel problem, we develop three different trainable models based on RNN and Generative Adversarial Network (GAN). The models (bucket, filter bank, and end-to-end) differ in how much expert knowledge is encoded, with the most general version being purely end-to-end. To train these systems, we use Amazon Mechanical Turk to collect textual descriptions for around 2000 image pairs sampled from several datasets. Experimental results evaluated on our dataset validate our approaches. In addition, given that the filter bank model is a good compromise between generality and performance, we investigate it further by replacing RNN with Graph RNN, and show that Graph RNN improves performance. To the best of our knowledge, this is the first computational photography work on global image editing that is purely based on free-form textual instructions.  
	
\end{abstract}

\section{Introduction}
\label{intro}

Consumers are increasing relying on portable embedded devices such as smartphones and tablets for their everyday activities. These devices tend to have small form factors that preclude fine-grain spatial control using the display. Adding voice-based instruction (systems such as Siri, Cortana, and Alexa) significantly enhances the capabilities of such devices. An application that would significantly benefit is photo editing. 

With few exceptions, interactive photo editing systems are primarily manual and often require significant display real estate for the controls. To allow a photo editing system to be voice-controlled, the mapping of voice to text to invocation of image operations requires domain-specific conversion of text to APIs. One solution is to handcraft this conversion by manually defining rules for editing effects (as was done in \cite{pixeltone}). However, this approach is hard to scale. 
	
\begin{figure}[H]
\centering
\includegraphics[width=0.75\textwidth]{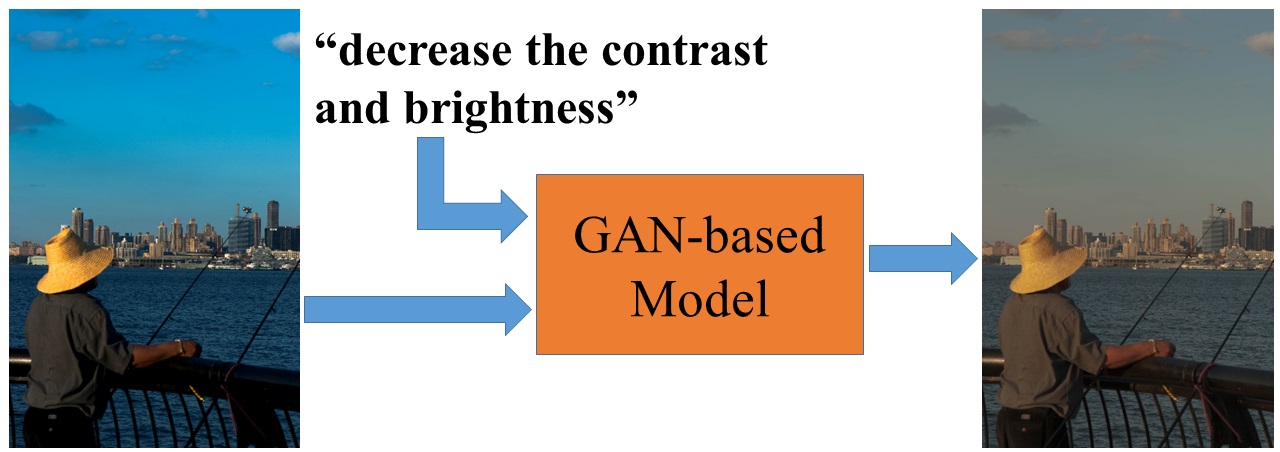}
\caption{Overview of our system. The inputs are an image and textual command, with the output being the result of applying the command to the input image.}
\label{fig:intro}
\end{figure}

\begin{figure}[H]
\centering
\includegraphics[width=0.9\textwidth]{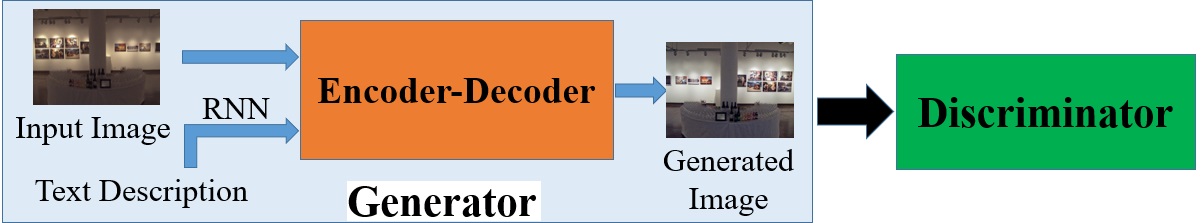}
\caption{Our GAN-based system.} 
\label{fig:overall}
\end{figure}

In this paper, we demonstrate global image editing through text, as illustrated in Figure~\ref{fig:intro}. Compared to other work \cite{imgsp,pixeltone}, our system is end-to-end trainable and easier to extend, since it does not require significant handcrafting of rules. We designed three different models based on Generative Adversarial Network (GAN)~\cite{GAN}. Our main contributions are:
\begin{itemize}[noitemsep,topsep=0pt]
	\item We believe our work is the first to tackle the general image editing problem under free-form text descriptions.
	\item We collected a database of image transformation pairs and their corresponding textual descriptions.
	\item We designed three different models: handcrafted bucket-based model, pure end-to-end model, and filter bank based model. Experimental results demonstrate the effectiveness of our approaches.
	\item We are the first method to apply graph RNN to text-image synthesis and demonstrate its effectiveness.   
\end{itemize}

In our work, we limit image editing to global transforms\footnote{Code is available at \href{https://github.com/sohuren/Img_edit_with_text}{https://github.com/sohuren/Img\_edit\_with\_text}. Supplementary file is online at author's homepage}. 

\section{Related work}
\label{relatedwork}

In this section, we briefly describe two voice-assisted systems, namely, PixelTone~\cite{pixeltone} for image editing, and Image Spirit~\cite{imgsp} for refining a parsed image. We also survey representative approaches for automatic image editing (specifically, image enhancement and style transfer), joint image-language analysis, and techniques that use attention or graph RNN.

\smallpar{PixelTone and Image Spirit.}
From application side, PixelTone~\cite{pixeltone} is the system most related to ours. It allows the user to edit the image through the voice command such as ``change the t-shirt to blue", after the t-shirt region is tagged. The system contains a speech recognition engine, a text analysis module, and an execution module. After converting the user's voice command to text, the text analysis module produces the atomic operation which can be run by the execution module. The text analysis module is based on predefined rules and NLP techniques such as tokenization and part of speech tagging. One limitation is that the predefined rules are manually constructed.   

Image Spirit~\cite{imgsp} is a system that parses an image into regions with semantic labels, and allows the user to verbally refine the result. Typical verbal commands include correcting an object label and refining a specific label. Based on the initial image parsing result, Image Spirit updates the local relationship between different objects in an MRF~\cite{mrf} in response to the utterance input, resulting in the enhanced result. As with PixelTone, the commands are also predefined, and the scenario of refining the image parsing result is different from our image editing scenario.        

Unlike PixelTone and Image Spirit, our approach does not rely on pre-define commands or rules, rather, it learns an end-to-end model which takes arbitrary text and learns corresponding transformations, based on a corpus.

\smallpar{Image Manipulation with Language.}
Concurrent with our work, there are several techniques that address end-to-end trainable model for image manipulation with language~\cite{chen2017language,Seitaro2017,nlp4seg}. Chen et al.~\cite{chen2017language} developed attentive models capable of combining text and image to produce a new image. To extract meaningful information from text, they use the attention mechanism~\cite{luong2015effective}. They demonstrate two editing tasks of image segmentation and colorization with natural language; different losses are used for training the different tasks. The work of Seitaro et al.~\cite{Seitaro2017} is similar to Chen et al.~\cite{chen2017language}, but they only consider MNIST dataset with instructions related to position moving such as ``moving 6 to the bottom." By creating the artificial dataset, they explored what the model can learn; as a side effect, training on an artificial dataset limits practicality.

By comparison, our model focuses on general textual instructions, which makes it extensible to different kinds of instructions. Further, instead of using attention mechanism~\cite{luong2015effective,chen2017language}, we use graph RNN~\cite{peng2017cross}. Finally, our model is trained on real dataset collected from Amazon Mechanical Turk. 

\smallpar{Automatic Image Editing: Enhancement and Style Transfer.}
There are a number of approaches for automatic image enhancement. Machine learning techniques have been used to train on original-enhanced image pair databases for enhancing images~\cite{Vladimir,S.Hwang,A.Kapoor,jianzhou}. The approach of \cite{Zhicheng} is based on a trained deep neural network to predict the enhanced image. Another form of image editing is style transfer (exemplified by \cite{gatys,Youngbae,JoonYoungLee,Yiming}). Here, given one image and reference image, the goal is to generate the new image according to the reference image's style. The mapping is purely image-based. Unlike our work, all these techniques do not act on a textual description.

\smallpar{Joint Image and Language Analysis.}
A significant amount of work has been done on joint image-language analysis. Topics in this space include image caption generation \cite{st,minde,jeff,chenxi,bodai}, video story telling \cite{Ting,Venugopalan}, visual question answering \cite{antol,yezhou,Nasrin,Das1}, image retrieval under natural language \cite{nlp4image}, object retrieval under language \cite{nlor,gtp,nlp4seg}, image synthesis from text \cite{GATIS,att2img,xu2017attngan,hong2018inferring} and referring expression generation \cite{gcuod,mohit2017,cvpr2017}. 

The topics of image retrieval under natural language, object retrieval, image synthesis from text, and referring expression generation are most relevant to our work. Ulyanov et al.~\cite{nlp4image} use natural language to guide image retrieval; image-text correspondence is used to find a common embedding space. There are techniques that, given text and image, localize a target object as a bounding box~\cite{nlor,gtp} or segment~\cite{nlp4seg} within the image. The work of Mirza and Osindero~\cite{gcuod} addresses the problem of referring expression generation, i.e., given image and a bounding box, generate the expression that can describe it. The techniques of \cite{cvpr2017} and \cite{mohit2017} generalize this problem in the context of reinforcement learning.

Given image objects and text, the approaches of \cite{inlpali} and \cite{dvsa} find the alignment between them. Kong et al.~\cite{ttico} find the alignment between text and RGB-D image, and use the text description to guide 3D semantic parsing. They show that image information helps to improve the language analysis result.

In \cite{GATIS,att2img}, the output image is synthesized from noise vector and text description, but in our work, we begin from the original image and try to transform the image under text description. The technique of \cite{GATIS} generates a fixed size image while our output image size depends on the input size, which complicates the image generation problem. Additionally, we rely on basic image concepts such as saturation and brightness while the techniques of \cite{GATIS,att2img,xu2017attngan} analyzes image content (with only image concept involved being color).    

In summary, all these techniques that focus on joint image-language analysis are not designed to transform the image using text. However, if we were to transform the image locally, as in ``change the color of the dog on right to white'', we would need to first localize the dog before applying the color change. Here, techniques such as \cite{nlor,gtp,nlp4seg} would be good candidate components to add to our system.

\smallpar{Attention and graph RNN.}

Attention has been used in various joint image and text problems, and generally there are two different attention mechanism: attention between different tokens in text~\cite{chen2017language}, and attention between tokens in text and pixels in image~\cite{chenxi,xu2017attngan}. 

Graph RNN is first used for cross-sentence $N$-ary relation extraction~\cite{peng2017cross}, and it subsumes plain\&tree RNN~\cite{tai2015improved}. Briefly, a graph RNN generalizes a linear-chain RNN by incorporating arbitrary long-ranged dependencies besides word adjacency. A word might have precedents other than the prior word, and its LSTM unit is expanded to include one forget gate for each precedent. For efficient training, a graph is decomposed into a forward pass and a backward pass, each consisting of edges pointing forward and backward, respectively. Backpropagation is then conducted on these two directed acyclic graphs, similarly to BiLSTM. (If a graph LSTM contains no edges other than word adjacency, it reduces to BiLSTM.) Additional dependencies include syntactic dependencies, discourse relations, coreference, and connections between roots of adjacent sentences~\cite{peng2017cross}.

In our work, we handle only global editing; given our limited training data, how to extract meaningful semantics from text is crucial. As such, graph RNN might be a more natural choice than attention since they can utilize graph structure provided by parser~\cite{manning2014stanford}. To the best of our knowledge, this is the first time that graph structured RNN is used in a joint text and image analysis problem.

\section{Model}
\label{model}

Our goal is to use text and an image as input and generate a new image globally transformed under the text description. This problem is well-suited to the adversarial framework provided by Generative Adversarial Network~\cite{GAN,c_gan}. The GAN objective function is a min-max problem, which is typically optimized in an alternating manner: 
\begin{align}
	\min_{\theta_{G}} \max_{\theta_{D}} \  ( &\mathbb{E}_{x \sim p_{data}(x)} [\log D_{\theta_{D}}(x)]  + \mathbb{E}_{z \sim p_{z}(z)} [\log ( 1-D_{\theta_{D}} (G_{\theta_{G} }(z) ) )] ),
\end{align}
where $D_{\theta_{D}}$ is the discriminator with parameter $\theta_{D}$ and $G_{\theta_{G}}$ is the generator with parameter $\theta_{G}$. The generator tries to confuse the discriminator while the discriminator differentiates between samples from true data distribution $p_{data}$ and samples from the generator given noise data distribution $p_{z}(z)$. 

In our work, the image transformation is achieved by the generator, which consists of an encoder-decoder architecture and a Recurrent Neural Network (RNN). As for the discriminator, $p_{z}(z)$ is the original image while $p_{data}$ is the corresponding edited image. The system is depicted in Figure~\ref{fig:overall}.

We design three models, and each model handles the text information differently:

\begin{enumerate}
	\item Hand-crafted bucket-based model, where similar image transformations are grouped prior to training as buckets. Each bucket has its own encoder-decoder architecture. 
	\item End-to-end model, with a single encoder-decoder architecture to handle the image and an RNN to handle text.
	\item Filter-bank model, where transformations are specified as trained convolution filters.
\end{enumerate}

All these models have exactly the same discriminator, and they only differ in the generator.

\subsection{Discriminator}
\label{discriminator}

We describe the discriminator first since it is same for all three models. We consider the conditional GAN (c-GAN)~\cite{c_gan} where the loss is also conditioned on the input image. However, compared with \cite{c_gan}, our discriminator also need to be text aware because the image is enhanced under the corresponding text description. 

Our discriminator uses four inputs: original image $I_{input}$, ground truth image $I_{gt}$ or generated image $I_{g}$, and the corresponding text $T_{des}$. As with \cite{GATIS}, for each image pair, we also consider sampling random texts (see in supplementary) to make the discriminator more text-aware.

Let $h(x)$ be an encoding function (e.g., an RNN) which can encode text $x$ into a vector. We first encode the text and down-sample the image before we depth-concatenate $I_{input}$, $I_{g}$, and $h(T_{des})$, then we feed the resulting vector to the discriminator with a negative label. 
In contrast, we feed the triple $I_{input}$, $I_{gt}$ and $h(T_{des})$ to the discriminator with a positive label. 

Additionally, the triple $I_{input}$, $I_{gt}$ and $h(T_{random})$ and $I_{input}$, $I_{g}$ and $h(T_{random})$ are treated as negative instances. The discriminator loss is just summed over all instances.

\begin{figure}[H]
\centering    
	\includegraphics[width=0.8\textwidth]{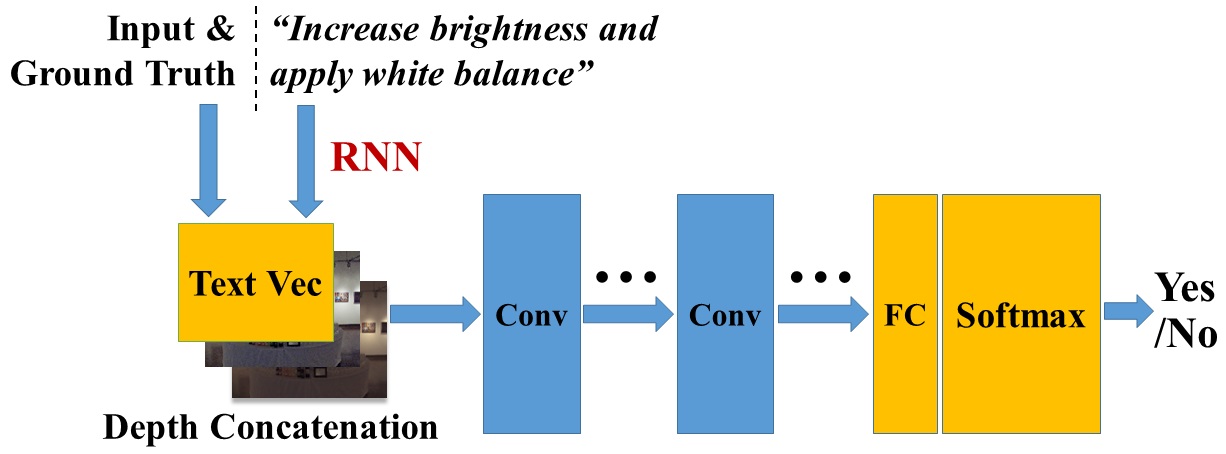}
	\caption{Our discriminator architecture.}
	\label{fig:discriminator}
\end{figure}

\begin{figure}[H]
\centering	
	\includegraphics[width=\textwidth]{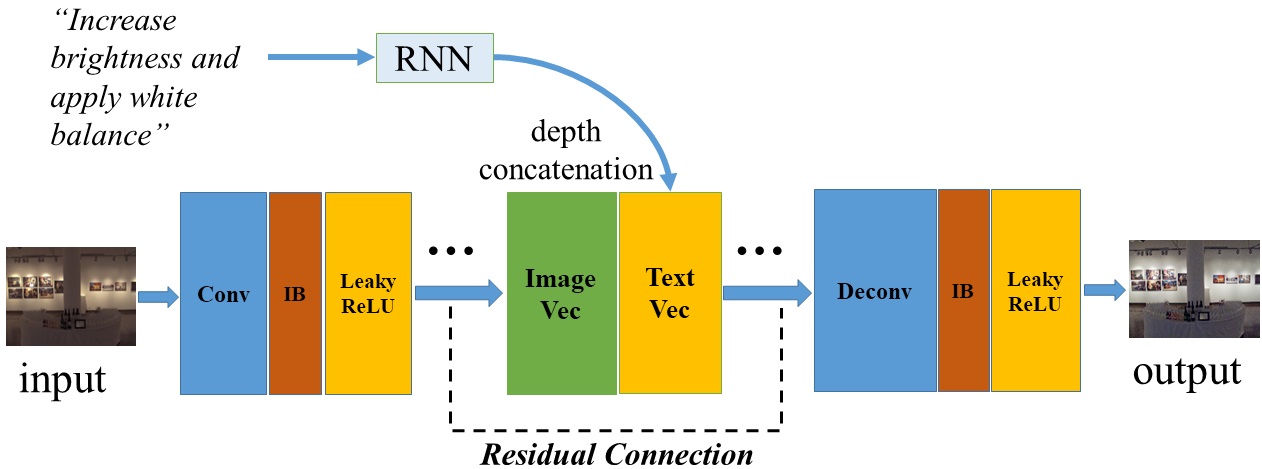}
	\caption{Our end-to-end model.}
	\label{fig:end-to-end}
\end{figure}

The discriminator architecture is depicted in Figure~\ref{fig:discriminator}. It is has fewer layers, and each layer contains convolution, instance normalization~\cite{instancenorm}, and activation function.
Compared with \cite{GATIS}, we use the sampled text in the context of c-GAN while \cite{GATIS} use the sampled text in basic GAN. More details on the discriminator are provided in the supplementary file.

\subsection{Generators}
\label{transformation_text}

All three generators take an image and text as input and generate a corresponding transformed image. Before describing our three models, we define the loss function. Given original image $I_{input}$, generated image $I_{g}$ and ground truth $I_{gt}$, we use the following losses to train the generator: 
\begin{itemize}[noitemsep,topsep=0pt]
	\item \textbf{Content loss}: 
	\begin{equation}
		l_{\text{content}} = \frac{1}{WHC}\sum_{c=1}^{C}|I_{g}^{c}-I_{gt}^{c}|_{1} ,
	\end{equation}
	where $W$, $H$, and $C$ are the image width, image height, and channel number, respectively. $l_{\text{content}}$ measures $l_{1}$ loss of the generated and ground truth images.
	\item \textbf{Adversarial loss}:   
	\begin{equation}
		l_{\text{adversarial}} = 1 - \log (D_{\theta_{D}}(I_{input}, I_{g}, h(T_{des}))) ,
	\end{equation}
	where $h(T_{des})$ is defined in (\ref{eq:Tdes}).
	$l_{\text{adversarial}}$ comes from the discriminator; it measures the similarity of the generated image with respect to the ground truth, conditioned on the input image. By minimizing it, the generator tries to fool the discriminator.  
	\item \textbf{Perceptual loss}: 
	\begin{equation}
		l_{\text{perceptual}} = \frac{1}{L}||F_{vgg\_19}(I_{gt}) - F_{vgg\_19}(I_{g})||_{2} ,
	\end{equation} 
	where $F_{vgg\_19}(I) = \text{Concat}(RELU_{2}(I), RELU_{3}(I), RELU_{4}(I))$, $RELU_{i}(I)$ is the feature after $RELU$ activation function~\cite{Relu} in $i$th layer in VGG-19 network for image $I$, and $L$ is the length of the concatenated feature.
	As with \cite{Perceptual_feifei,Photo_Realistic}, we use the pre-trained VGG-19 network~\cite{vgg} to extract the high-level visual feature.
\end{itemize}

The final loss for the generator is a weighted combination of those three losses:
\begin{equation}
	l_{G} = l_{\text{content}} + \alpha l_{\text{adversarial}} + \beta \ l_{\text{perceptual}}.
\end{equation}
where $\alpha=1$ and $\beta=0.02$ based on tuning the validation set.

\begin{figure}[H]
\centering
	\includegraphics[width=0.7\textwidth]{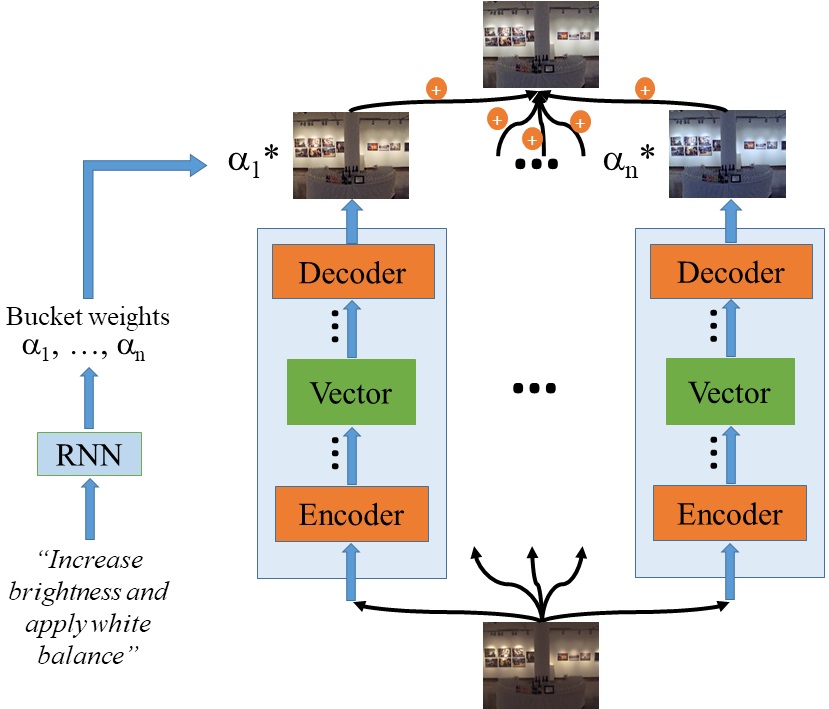}
	\caption{Bucket model.}
	\label{fig:bucket}
\end{figure}

\begin{figure}[H]
\centering
	\centering
	\includegraphics[width=0.7\textwidth]{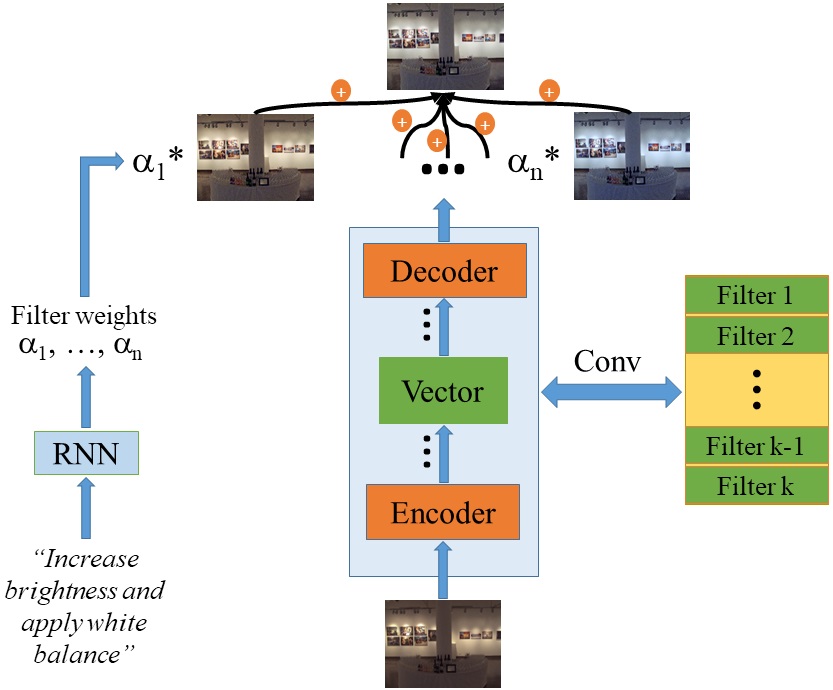}
	\caption{Filter bank model.}
	\label{fig:filterbank}
\end{figure}

\smallpar{Bucket Model.} One design is the bucket model, which is based on the idea that similar image transformations should be grouped as buckets. Each bucket represents a different image transformation (e.g., one for increasing the brightness, another for reducing the contrast). The disadvantage is the grouping is manual. The architecture of the bucket model is shown in Figure~\ref{fig:bucket}. 

Given some text, we train an RNN to generate a distribution over buckets, and the final generated image is a weighted linear combination of different buckets. Let
\begin{equation}
	h(T_{des}) = \overrightarrow{RNN}(t_{1}, \ldots ,t_{n}) || \overleftarrow{RNN}(t_{1}, \ldots ,t_{n}) ,
	\label{eq:Tdes}
\end{equation}
where $\overrightarrow{RNN}$ and $\overleftarrow{RNN}$ are the last hidden state vectors when text is fed to RNN in opposite directions.
Let weight $\alpha = \text{softmax}(h(T_{des}))$, with $K_b$ buckets and the output of each bucket being $I_{k}$. (In our work, $K_b = 5$.)
The final output image $I_{g}$ is a weighted linear combination from the different buckets, i.e., $I_{g} = \sum_{k=1}^{k=K} \alpha_{k}I_{k}$.

In this model, each bucket has its own encoder-decoder architecture. The encoder is a down-sampling procedure which contains a series of conv-batch normalization~\cite{batchnorm,instancenorm} Leaky ReLU units~\cite{Relu}. The decoder is similarly constructed, except in reverse order to constitute an up-sampling procedure. 
In our implementation, the down-sampling and up-sampling networks have the same depth, and optionally we can use skip connection~\cite{resnet}, i.e., we concatenate the $i$th layer in down-sampling network with the $(N-i)$th layer in up-sampling network.

To group the image, several methods can be used: surface form level word matching, cluster over word (sentence) embedding, or manually design the buckets. In our work, however, we manually designed the buckets based on the bigram distribution shown in the supplementary file. We have tried using automatic grouping methods, but they appear to be less effective.

\smallpar{End-to-End Model.} The bucket model, while straightforward, requires some handcrafting of the buckets. Inspired by \cite{GATIS}, we also design an end-to-end model. We use another RNN (which is different from that used in the discriminator) to encode the text to a vector; this vector is then concatenated with the image vector. 
As with the bucket model, we also use the encoder-decoder framework to encode the image. 
The overall architecture of the end-to-end model is shown in Figure~\ref{fig:end-to-end}.

Given an image $I_{img}$, we first encode it through a deep convolution neural network as $\text{Encode}(I_{img})$, followed by a depth concatenation between this image vector and the text vector:    
\begin{equation}
	h(T_{des}, I_{img}) = \text{DepthConcat} (\text{Encode}(I_{img}), h(T_{des} )) .
\end{equation}
Subsequently, we feed $h(T_{des}, I_{img})$ to the decoder. 

\smallpar{Filter Bank Model.} An end-to-end model is conceptually elegant. However, making it work is difficult due to limited expressive power, especially if we consider that image transformations can be bidirectional. An example is with respect to brightness, where the user can specify to ``increase the brightness" or ``decrease the brightness". The bucket model is easy to understand, but it requires pre-designing the buckets; incorporating additional data will likely to require changes to the bucket design. Our third model, the filter bank model, is designed to combine the advantages of these two models. The architecture for the filter bank model is depicted in Figure~\ref{fig:filterbank}.

Given a description, an RNN is used to encode it and generate a distribution over different filters, which is conceptually the same as the bucket model. Each filter $F_{k}$ is a $k \times k \times c_{in} \times c_{out}$ convolution filter, and the final image is a weighted linear combination of different images.

Given an image $I_{img}$, to generate the enhanced image based on filter $F_{k}$, we first use the encoder to encode the image as a hidden vector. We then convolve this hidden vector with filter $F_{k}$, and the result is fed to the decoder. With $K_f$ filters, we have $K_f$ different images generated. (In our work, $K_f = 5$.) For the $k$th filter, we have
\begin{equation}
	I_{k} = \text{Decoder}(\text{Conv}(\text{Encode}(I_{img}), F_{k})) .
\end{equation} 
The final output image $I_{g}$ is obtained using $I_{g} = \sum_{k=1}^{k=K} \alpha_{k}I_{k}$. 
This model is similar to that described in \cite{stylebank}, but there is a major difference: The model in \cite{stylebank} is used for style transfer and each filter corresponds to one pre-determined style. During training, each training instance contains an image pair and corresponding filter id, and it only optimizes the corresponding filter and the shared encoder-decoder parameter. By comparison, for our filter bank model, the filters are jointly trained automatically from image pairs and the model learns how to decompose the transformation automatically (the only manual step is specifying the number of filters).     

\section{Data Collection}
\label{datacollect}

To train our models, we need original-edited image pairs with associated text descriptions. To the best of our knowledge, there is no such existing dataset. The MIT-Adobe 5k dataset~\cite{Vladimir} consists of original-edited image pairs generated by five professional photographers, but it does not contain text that describe the image transformation (such as brightness change and color balance). For each image pair, the list of operations used to generate the edited image is given; an operation consists of a software editing command and its associated parameters. The fine granularity of information is not useful for associating general casual description of the image transformation with the original-edited image pair. Other publicly available text-image datasets such as MS-COCO \cite{ms_coco}, ReferIt~\cite{referit}, and Flickr30k Entities~\cite{ijcv2016} contain text that describe the image content, but such text are not related to image editing or style. In addition, these datasets do not contain edited versions of the original.

As such, we ran a user study to collect our own dataset through Amazon Mechanical Turk. We use a random subset of the MIT-Adobe 5k dataset; for a given original-edited image pair, we ask the subject to type in a phrase to describe the image transformation. We also flip the order of the image pair to sample the reverse transformation.

Each task (``hit" in AMT parlance) involves describing transformations for 8 pairs. For each image pair, the subject was asked to rate the image transformation and describe the image operations that are applied to the original to produce the edited version.  

Procuring reliable data from such a user study is difficult because most Turkers are not highly familiar with concepts of photography, and as such, have only rudimentary vocabularies to describe visual changes. Initially, to assist with the task, we provided several example image pairs with plausible responses as guidelines. This unfortunately resulted in subjects copying and pasting example responses regardless of relevance. Even if they do not copy and paste responses, many users are not familiar with imaging concepts and provided inappropriate text.

In response to these issues, we made the following changes: 
(1) disabled copy and paste, 
(2) added examples (with explanations) that would cause their work to be rejected,
(3) added a qualification test to see if the subject understands color and contrast, and
(4) used heuristics to manually filter out ``bad" responses.
The new data are significantly better than those obtained through the trial run. By disabling cut-and-paste, the responses are much more varied. By explaining why responses may be rejected and enforcing a qualification test, data noise is significantly reduced.

The interface with examples is shown in Figure~\ref{fig:example}. (See in the supplementary file for additional examples, the qualification test, and task interface.)

\begin{figure}[H]
\centering
	\includegraphics[width=0.8\textwidth]{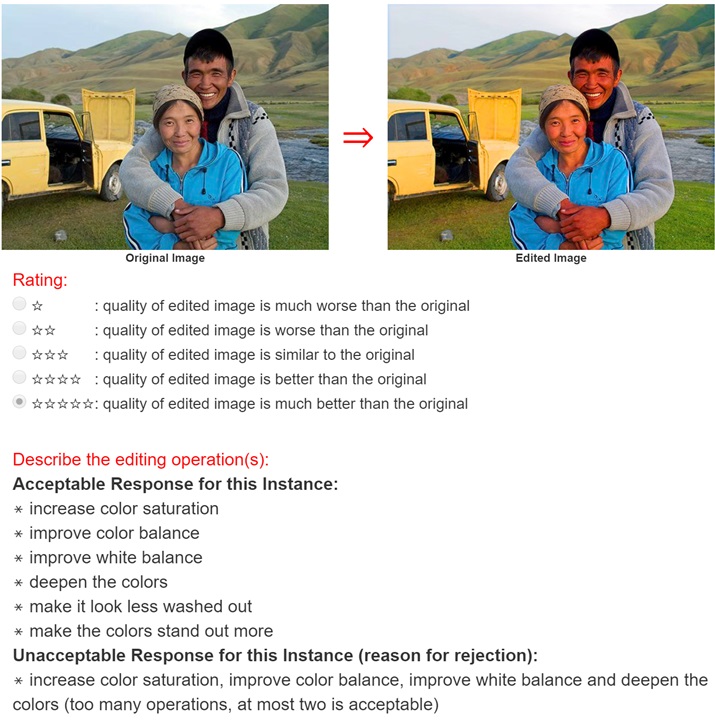}
	\caption{Guidelines and example responses provided in the user study.}
	\label{fig:example}
\end{figure}

\begin{figure}[H]
\centering
		\begin{tabular}{ | c | c |}
			\hline
			{\bf Model} & {\bf $p$-value}  \\ \hline
			End2end  vs. GT  &  $7.4 \times 10^{-6}$\\ \hline
			GT vs. Bucket(a) & 0.007   \\ \hline
			End2end  vs. Bucket(f) & 0.02   \\ \hline
			GT  vs. Bucket(f) &  0.02  \\ \hline
			End2end  vs. Bucket(a) & 0.06  \\ \hline
			FB vs. End2end & 0.09  \\ \hline
			FB vs. GT  &  0.09 \\ \hline
			FB vs. Bucket(f) & 0.53   \\ \hline
			FB vs. Bucket(a) & 0.58   \\ \hline		 
			Bucket(a) vs. Bucket(f) &  0.67  \\ \hline
		\end{tabular}
	\caption{Pairwise comparison between different models, the lower the value of $p$, the more different the two models are. }		
	\label{table:pairwise}
	
\end{figure}

Once the data have been collected, we further manually check the response. We removed responses that are obviously inconsistent with the actual image operation, too generic (e.g., ``beautify the image"), or are not descriptions (e.g., ``the edited image need to be brighten" in response to the edited image being a darkened version of the original).
Totally 370 responses are removed in this way, which count 15\% of all the raw responses.

We end up with 1884 image pairs and annotations, with each image pair having on average 1.6 text annotations. 1378 image pairs are used for training, 252 for validation and 252 for test. 
These image pairs contain multiple image transformation directions, e.g., improving the color balance, increase/decrease the image brightness, increase/reduce the image saturation, deepen the colors and keep the image the same. For additional statistics on the collected data, see the supplementary file.

\section{Implementation}

We use Pytorch to implement our models. We tried two different versions of encoder-decoder: one is a typical encoder-decoder without skip connection, with the other with skip connection~\cite{resnet,pix2pix}. We find the version with skip connection has better performance and faster training. We use Adam~\cite{Adam} as the optimizer, with the initial learning rate 0.001, and will half the learning rate if we don't observe the loss reduction on validation set.    

Our RNN is one layer bidirectional Gated Recurrent Unit (GRU)~\cite{gru} with a hidden size of 128. The vocabulary size is around 4k and word embedding size is 200. We initialize the word embedding with the pre-trained Glove word embedding~\cite{glove}.                 

For our bucket and filter bank models, due to memory constraints, we limit the numbers of buckets and filters to 5 each, i.e., $K_b = K_f = 5$. For our bucket model, we have $K_b$ encoder-decoder pairs without any shared parameters. As a result, the optimization process requires a large amount of memory; in addition, it is slow, especially during back propagation. To overcome this problem, we pre-train the $K_b$ independent encoder-decoder pairs separately and fix them when training the bucket model. Additionally, for the bucket model, after training, we compute two image outputs, one with the highest weight (``argmax", i.e., Bucket(a)) and another being a weighted average (``fusion", i.e., Bucket(f)). 

On the other hand, the filter bank and end-to-end models are trained from scratch, since their memory requirements are not as severe and the training is faster. Specifically, bucket model, including the parallel pre-training for different buckets, totally takes 40 hours and needs 4 GPUs, while filter bank takes 25 hours and only need 1 GPU, and the end-to-end model only needs 20 hours and 1 GPU.
More implementation details are given in the supplementary file.

\section{Experimental Results}
\label{result}

In this section, we first report results for automatic image enhancement (without text) as a sanity check. We then describe the results of a user study to evaluate the performance of our models in producing the edited image given an input image and text description. Finally, we show the effects of the trained filters from our filter bank model.

\subsection{Automatic Image Enhancement}

\begin{figure}

	\begin{subfigure}{.33\linewidth}
		\centering
		\includegraphics[width=.7\linewidth]{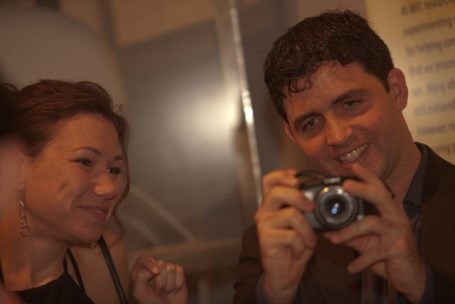}
		\label{fig:sfig1}
	\end{subfigure}%
	\begin{subfigure}{.33\linewidth}
		\centering
		\includegraphics[width=.7\linewidth]{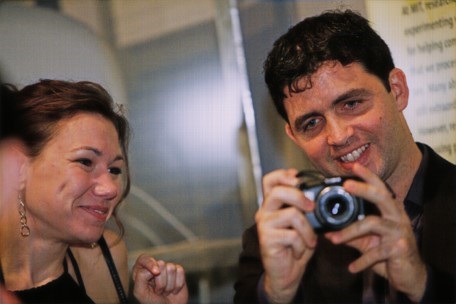}
		\label{fig:sfig2}
	\end{subfigure}
	\begin{subfigure}{.33\linewidth}
		\centering
		\includegraphics[width=.7\linewidth]{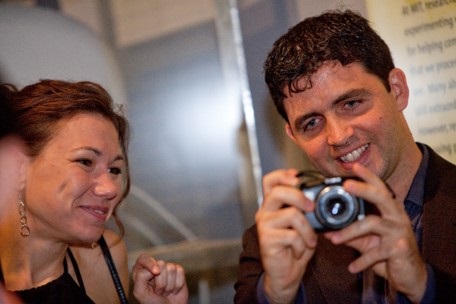}
		\label{fig:sfig3}
	\end{subfigure}	
	
		\begin{subfigure}{.33\linewidth}
		\centering
		\includegraphics[width=.7\linewidth]{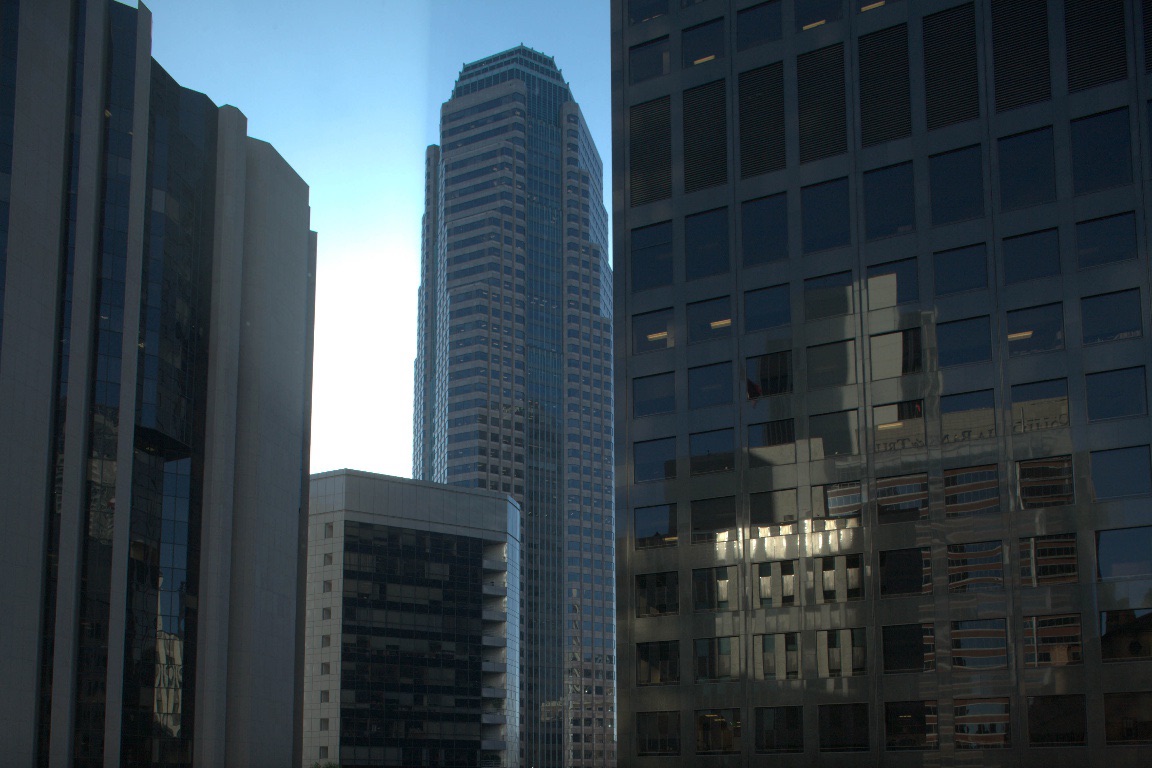}
		\caption{Input}
		\label{fig:sfig1}
	\end{subfigure}%
	\begin{subfigure}{.33\linewidth}
		\centering
		\includegraphics[width=.7\linewidth]{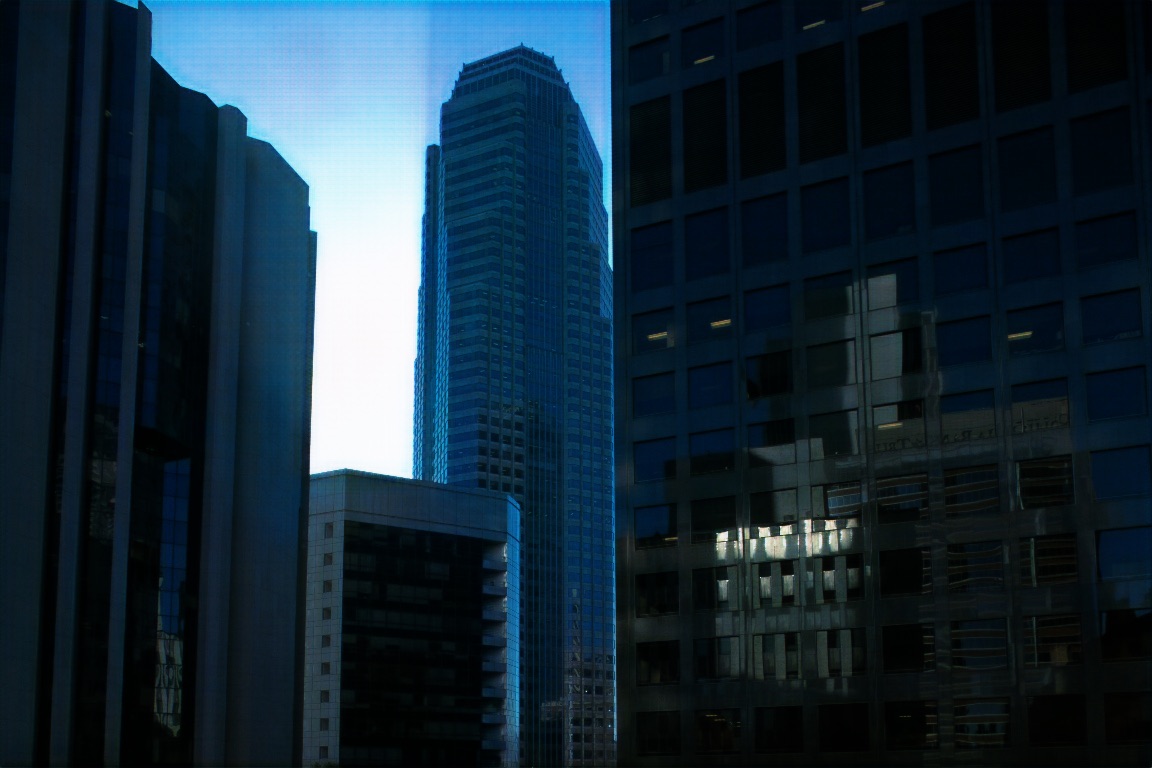}
		\caption{Output}
		\label{fig:sfig2}
	\end{subfigure}
	\begin{subfigure}{.33\linewidth}
		\centering
		\includegraphics[width=.7\linewidth]{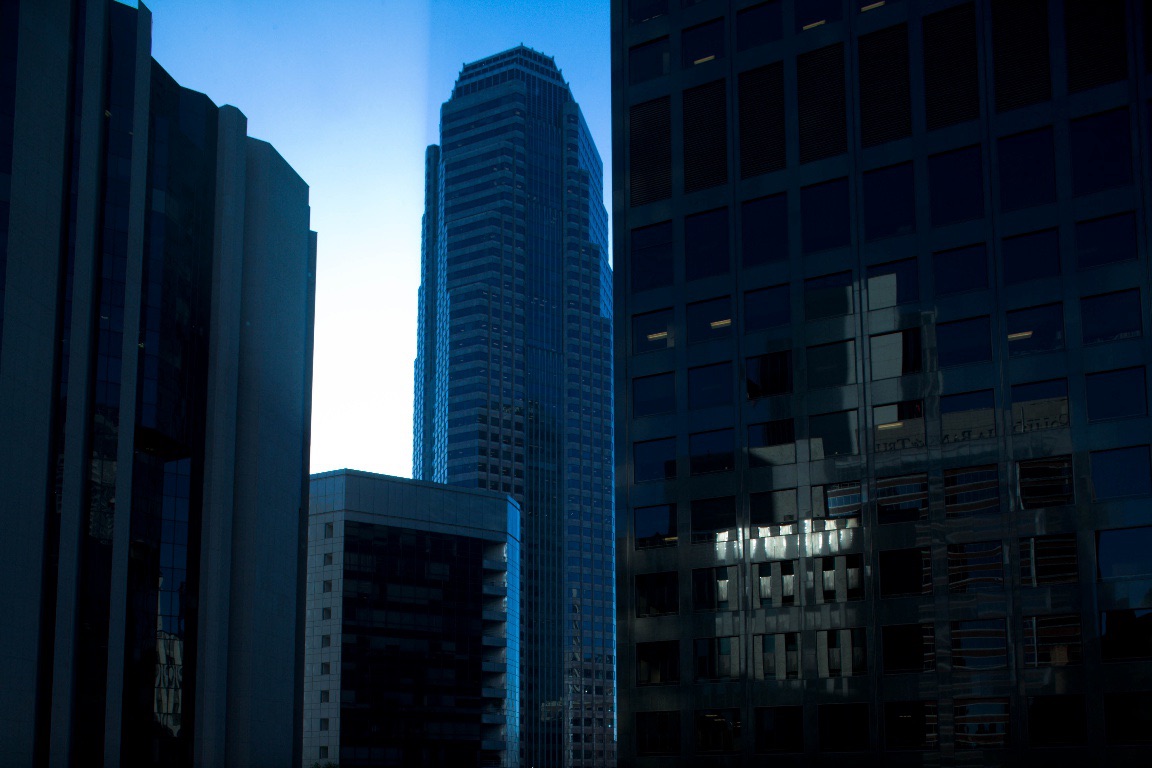}
		\caption{Ground Truth}
		\label{fig:sfig3}
	\end{subfigure}	
	
	\caption{Examples of automatic image enhancement.}
	\label{fig:autoenhancment}
\end{figure}

We first investigate the performance of c-GAN with encoder-decoder architecture in the context of automatic image enhancement, without any text used. We randomly selected 1200 image pairs (results from one expert) from MIT-Adobe fiveK and train the model; a representative result on the validation set is shown in Figure~\ref{fig:autoenhancment}. 
We quantitatively evaluate our automatic image enhancement performance. Table~\ref{table:score_auto} lists the L2 error (in L*ab space) for our method. Even though the randomly selected dataset in \cite{S.Hwang} is not the same with ours\footnote{The random dataset from \cite{S.Hwang} is not publicly available, so we instead randomly select the same number of images.}, but in general, we believe our results are representative. The results indicate that c-GAN with encoder-decoder as generator is suitable for our problem.  

\begin{table}
\caption{Comparisons of average L2 error on test sets, with standard error of 95\%.}
	\begin{center}
		\begin{tabular}{ | c | c |c|c|}
			\hline
			& Input & Hwang et al.~\cite{S.Hwang} & Ours \\ \hline
			Error & 17.1 $\pm$ 0.9 & 15.0 $\pm$ 0.8 & 12.1 $\pm$ 0.9   \\ \hline		
		\end{tabular}
	\end{center}
	\label{table:score_auto}
\end{table}

\subsection{Image Transformation from Text Description}

\begin{figure}
	\begin{subfigure}{.33\linewidth}
		\centering
		\includegraphics[width=.7\linewidth]{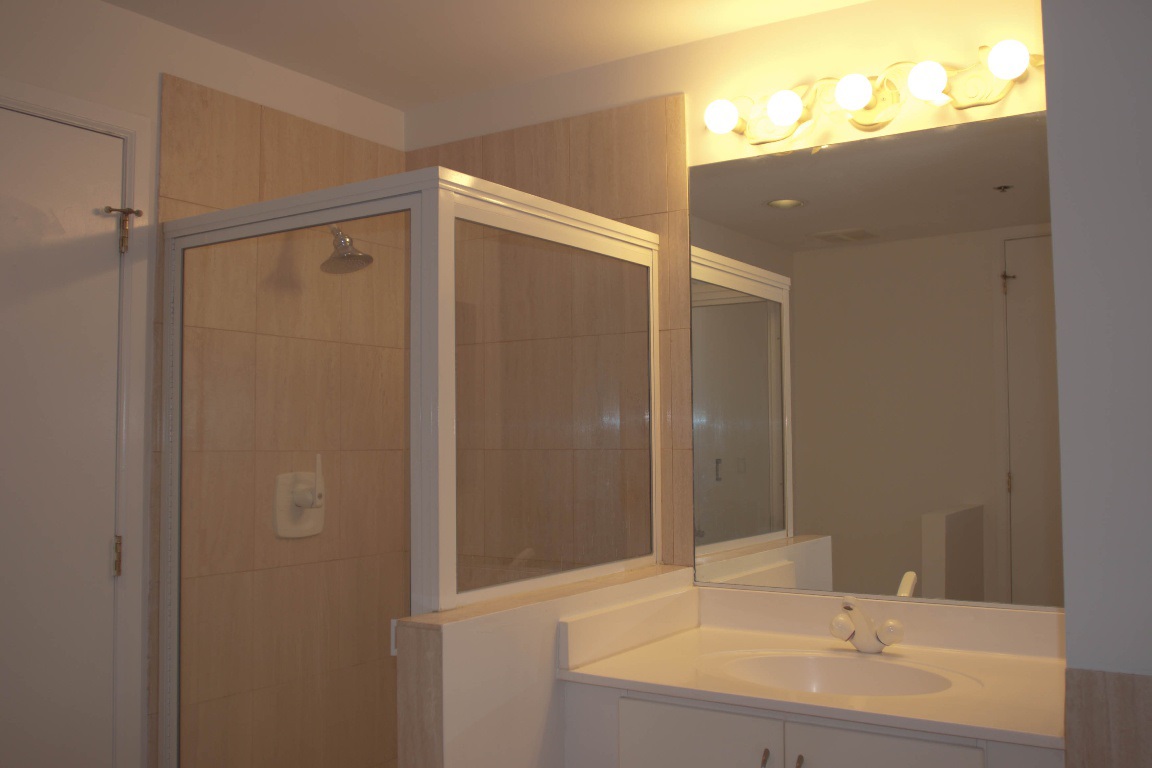}
		\caption{Input}
		\label{fig:sfig1}
	\end{subfigure}%
	\begin{subfigure}{.33\linewidth}
		\centering
		\includegraphics[width=.7\linewidth]{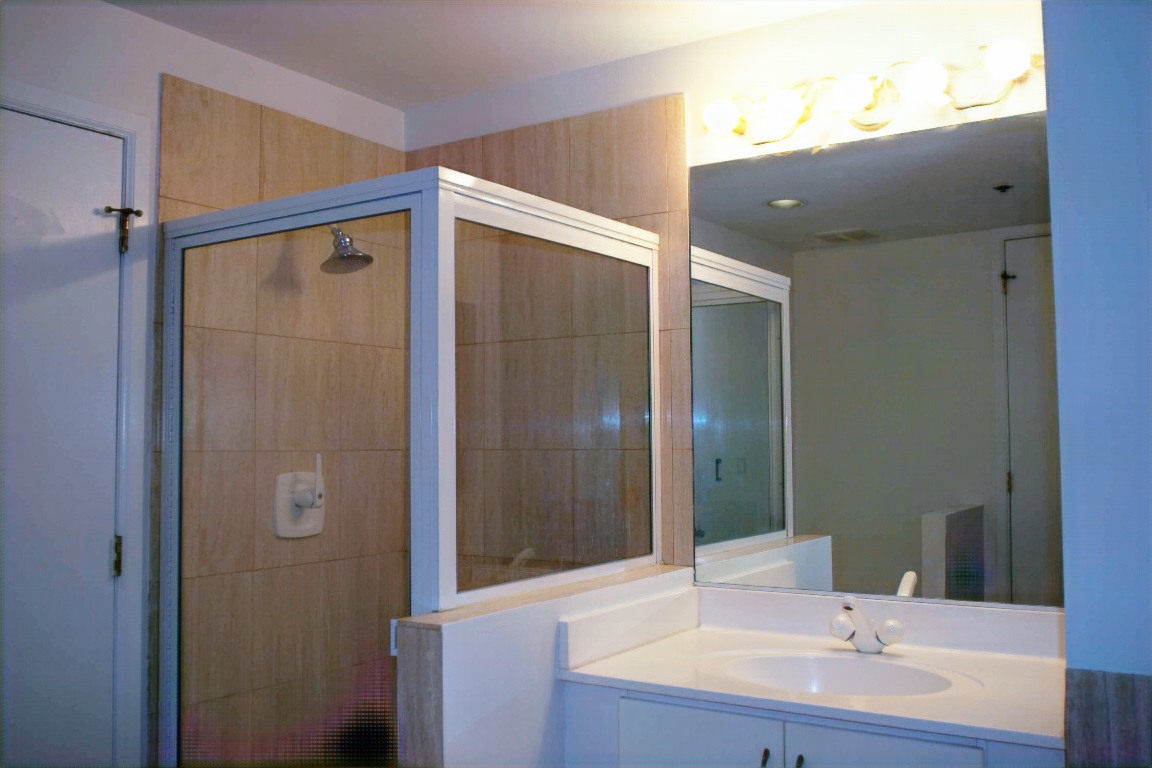}
		\caption{Output}
		\label{fig:sfig2}
	\end{subfigure}
	\begin{subfigure}{.33\linewidth}
		\centering
		\includegraphics[width=.7\linewidth]{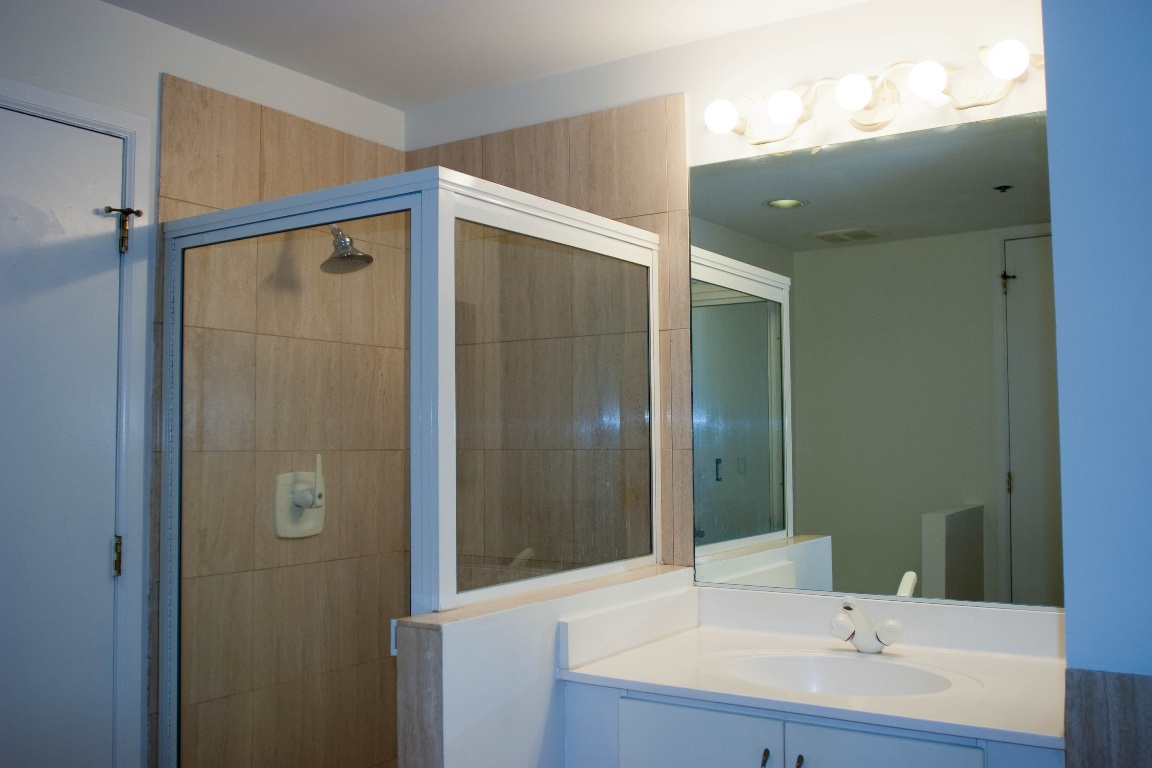}
		\caption{Ground Truth}
		\label{fig:sfig3}
	\end{subfigure}
	\caption{Example of image editing under textual description ``enhance white balance and contrast.'' More examples are in the supplementary material.}
	\label{fig:hancmentwithtext}
	\vspace{-0.2em} 
\end{figure}

Since there is no existing benchmark, we design a user study for such an evaluation. We are specifically interested in how well the edited image fit the text description given an input image, for all the models and ground truth\footnote{Please note that we are less interested in how measuring how close the generated edited image is to the ground truth, because such a metric takes the text description out of the loop.}. We want to extract metrics that are both absolute (through standalone rating) and relative (pairwise comparison). One representative result is given in Figure \ref{fig:hancmentwithtext}.  

\smallpar{Standalone rating:}
The subject is shown an original-edited image pair with text that describes the image transformation, and is asked to rate (on a scale of one to five stars) based on the instruction ``how well does the edited image follow the instructions?". There are five different pair versions, with the original image the same throughout and the edited image from ground truth, bucket model (fusion and argmax), filter bank model, and end-to-end model, respectively. 
The order of appearance is randomized. Each subject is shown eight image pairs corresponding to two different original images.
For this portion of the user study, each image pair get five ratings, and the rating for that pair is averaged.

\smallpar{Pairwise comparison:}
The subject is shown two image pairs as well as the text description, and is asked to pick the pair that fits the text better. The same four versions are used. Each subject makes eight comparisons. 

We obtained responses from 120 subjects; the reward for each task or ``hit" is US\$0.20.
(The user study interface is shown in the supplementary file.)
Results of the user study are listed in Tables~\ref{table:stat}.

\begin{table}
\centering
		\begin{tabular}{ | c | c | c |}
			\hline
			{\bf Model} & {\bf Mean} & {\bf Std Dev} \\ \hline
			Ground Truth  & 3.53 &  1.22  \\ \hline
			Bucket(f) & 3.36 & 1.25   \\ \hline
			Bucket(a) & 3.33 & 1.21   \\ \hline
			Filter Bank & 3.31 & 1.22   \\ \hline		
			End-to-End  & 3.19 &  1.30 \\ \hline
		\end{tabular}
	\caption{Standalone rating for the different models.}
	\label{table:stat}
\end{table}

\begin{figure}[H]
\centering
\includegraphics[width=0.9\textwidth]{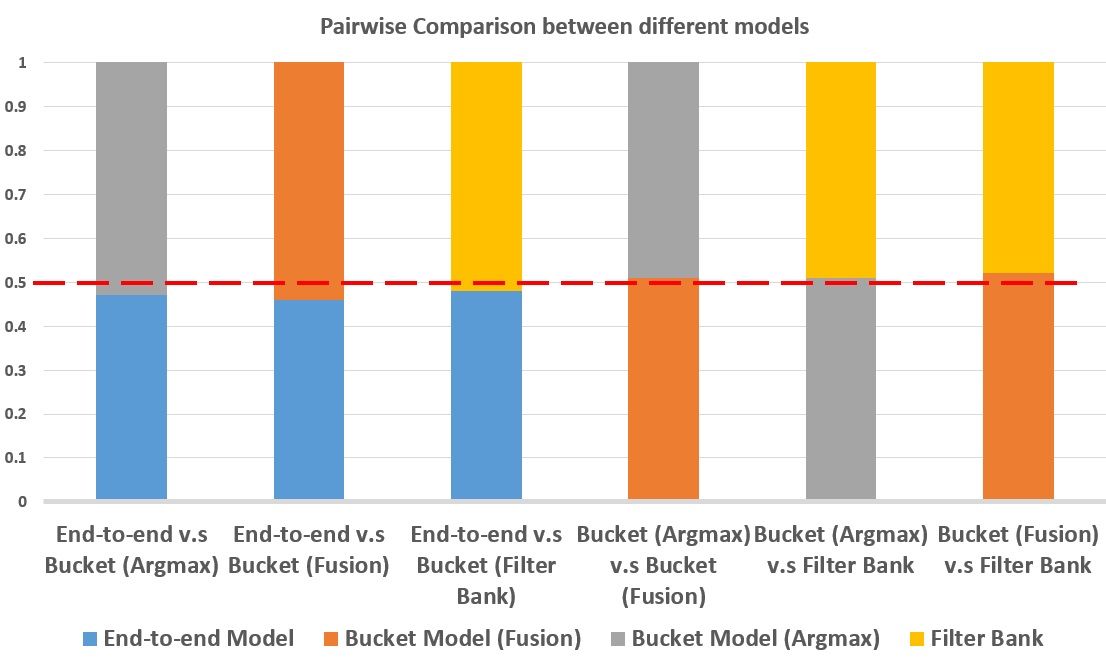}
	\caption{Pairwise rating between different models. Red dash line represents equal rating 0.5. }
	\label{fig:pairwise}
\end{figure}

Table~\ref{table:stat} shows that the bucket model has the highest rating among all the three models. This is not surprising since the bucket model is customized, with the disadvantage of being less scalable. The filter bank model is next best with the end-to-end model being third. While the end-to-end model is the most conceptually elegant with the least amount of user specification, it has only one encoder-decoder, which limits its expressive power. It is less able to learn multiple directional transformations. While the filter bank model also has only one encoder-decoder, it has filters between them; the filters can be interpreted as a type of bucket model with shared encoder-decoder parameters among the buckets.

Compared with the ground truth, however, the differences are not significant. Please note that the ground truth version has a score of only 3.53; this may be due to most users being not familiar with image concepts. 
Figure~\ref{fig:pairwise} shows pairwise ratings between different models, which is consistent with Table~\ref{table:stat}.

Table~\ref{table:pairwise} lists the $p$-values between the scores of different models, where smaller $p$-values implies larger difference between models\footnote{For a given model, we calculate the average score for each image pair and then evaluate the $p$-values between the scores of different models.}. Based on this table, the filter bank model is very close to the bucket model while the end-to-end model is less similar to the bucket model or filter bank model. 

From the practical point of view, the filter bank model appears to be the best choice since the performance is good while not requiring much manual effort (apart from selecting the number of filters). Additionally, it requires less memory than the bucket model. For same encoder-decoder architecture with $K_b$ buckets, it only need $1/K_b$ memory as that for the bucket model. In addition, the filter bank model does not require pre-training for different buckets, making it much more efficient. For the same amount of memory, the filter bank model can afford to incorporate more filters than there are buckets.  


\subsection{Effects of Automatically Trained Filters}
\label{filters_anaylysis}

\begin{figure}[ht!]
	\centering
	\begin{subfigure}{0.28\linewidth}
		\centering
		\includegraphics[width=0.7\linewidth]{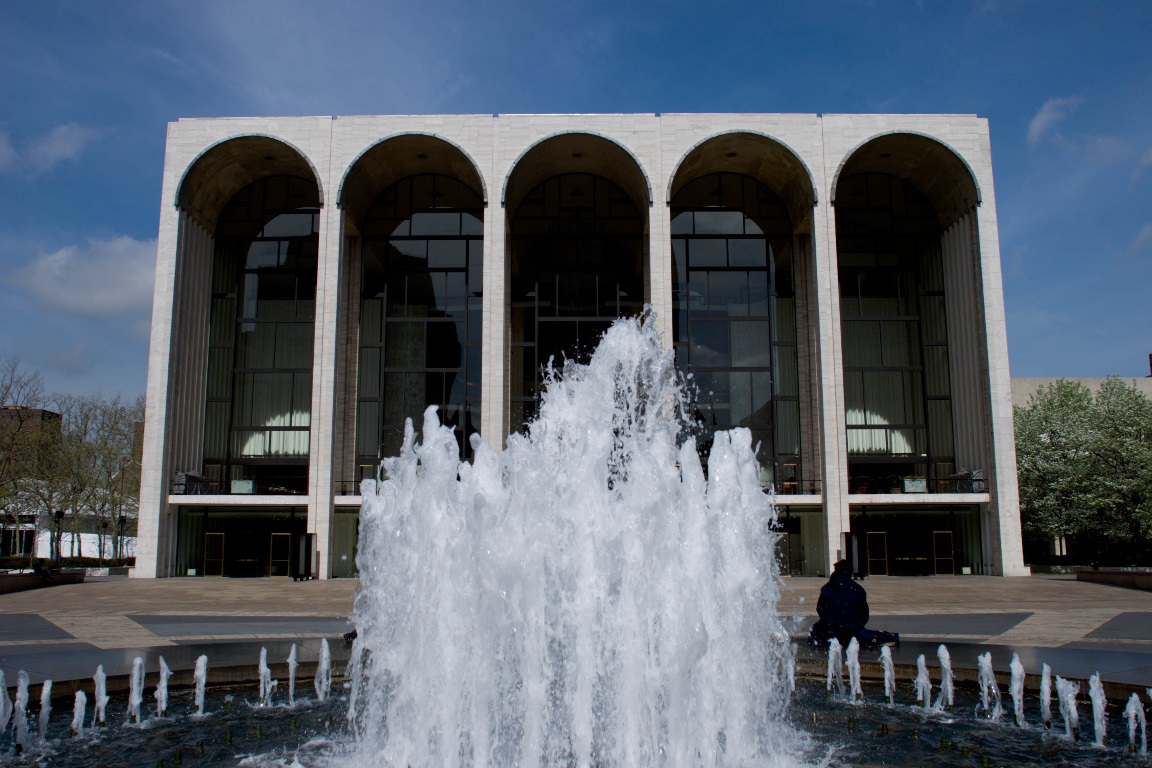}
		\caption{Input}
	\end{subfigure}
	\begin{subfigure}{.28\linewidth}
		\centering
		\includegraphics[width=0.7\linewidth]{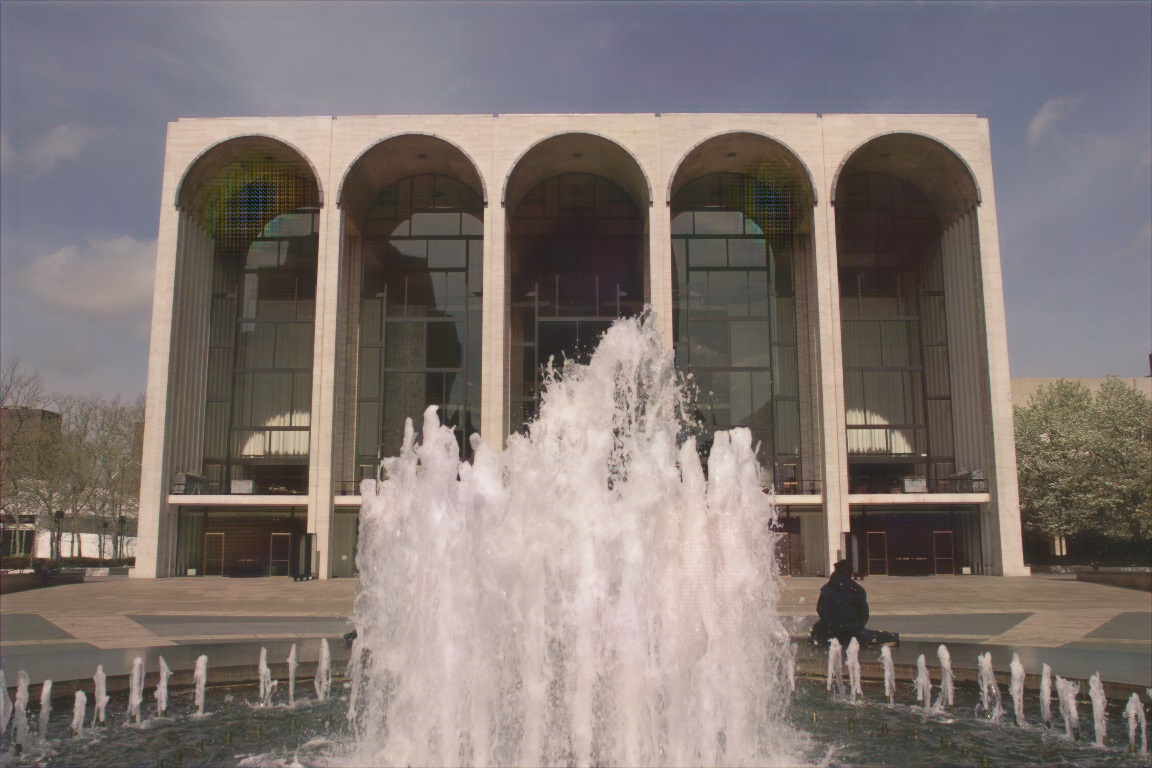}
		\caption{Using $F_{0}$}
	\end{subfigure}
	\begin{subfigure}{.28\linewidth}
		\centering
		\includegraphics[width=0.7\linewidth]{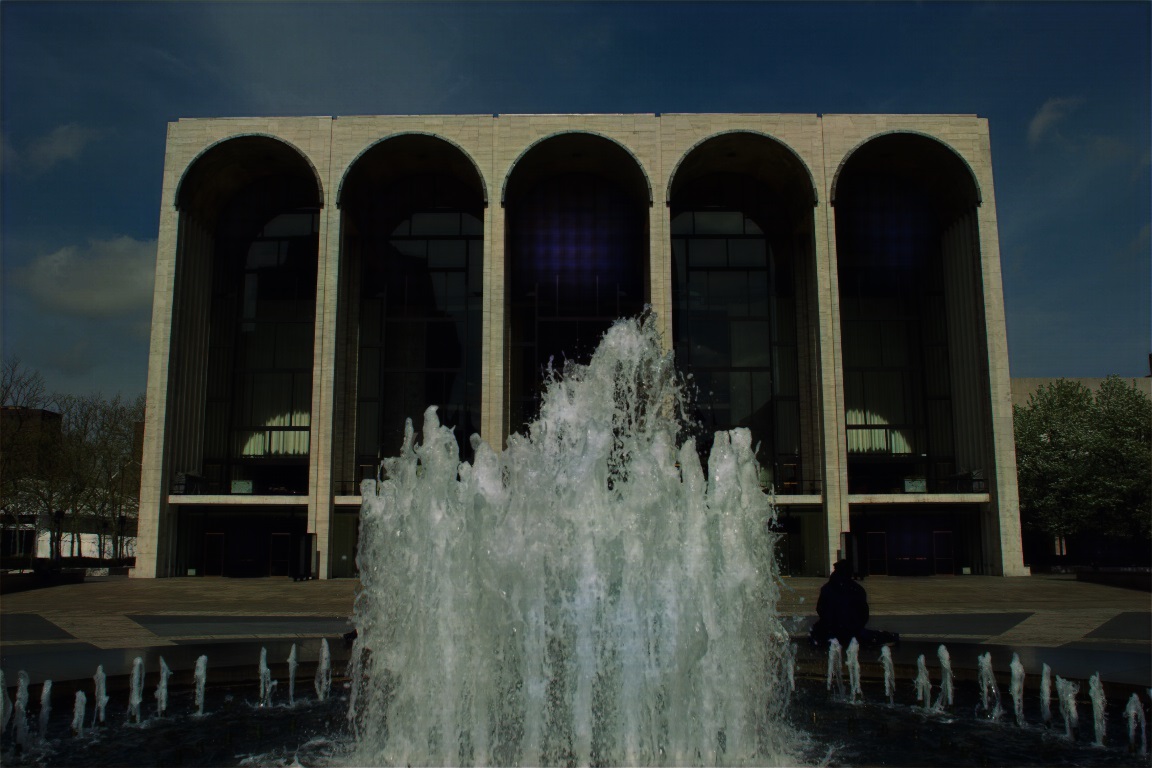}
		\caption{Using $F_{1}$}
	\end{subfigure}
	
	\begin{subfigure}{.28\linewidth}
		\centering
		\includegraphics[width=0.7\linewidth]{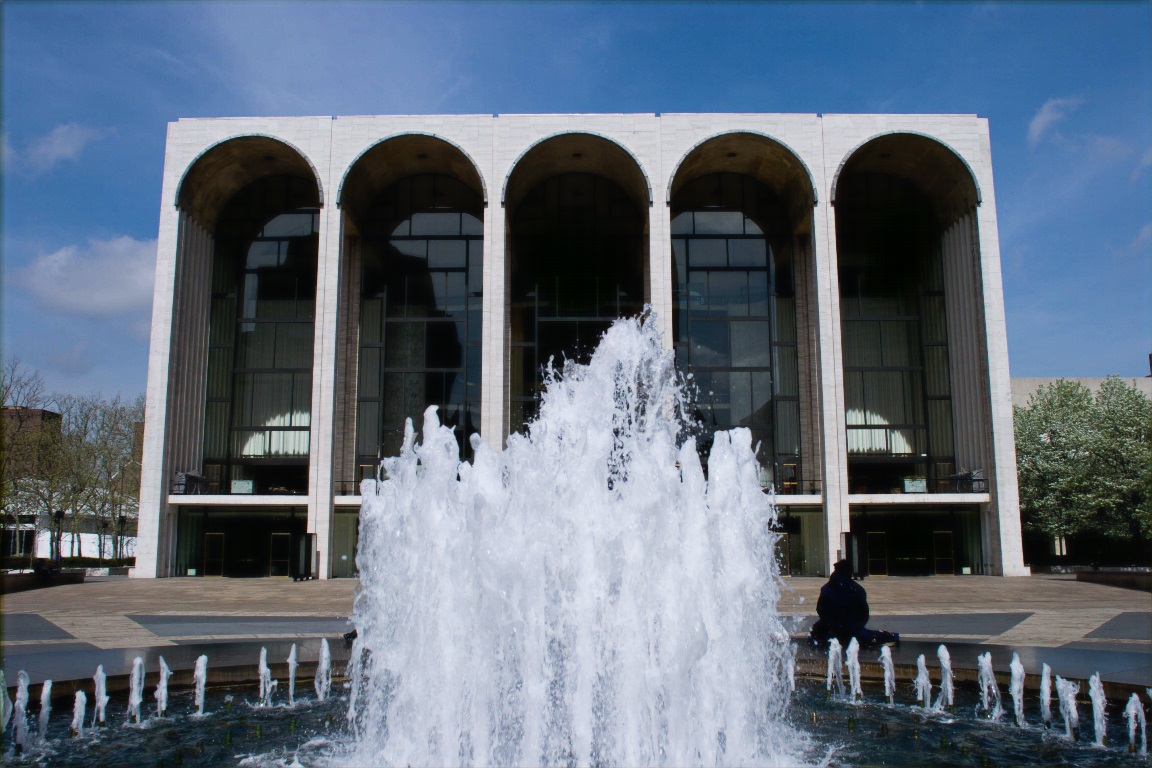}
		\caption{Using $F_{2}$}
	\end{subfigure}
	\begin{subfigure}{.28\linewidth}
		\centering
		\includegraphics[width=0.7\linewidth]{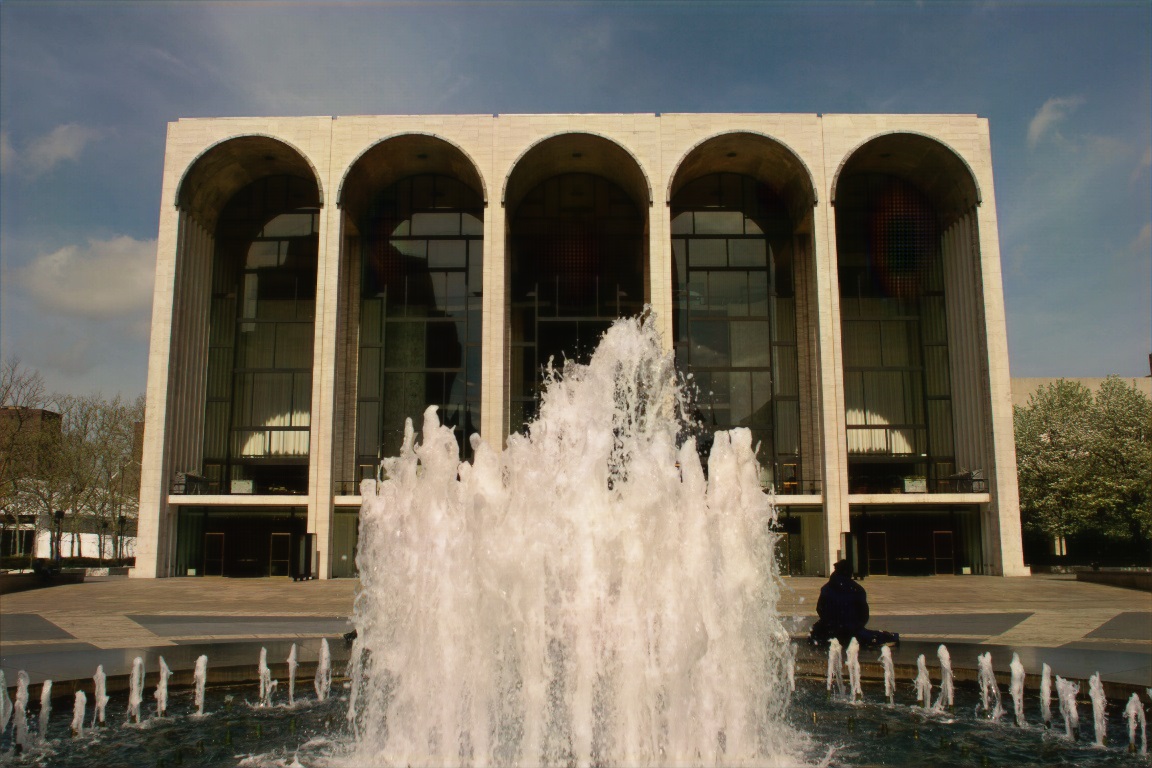}
		\caption{Using $F_{3}$}
	\end{subfigure}
	\begin{subfigure}{.28\linewidth}
		\centering
		\includegraphics[width=0.7\linewidth]{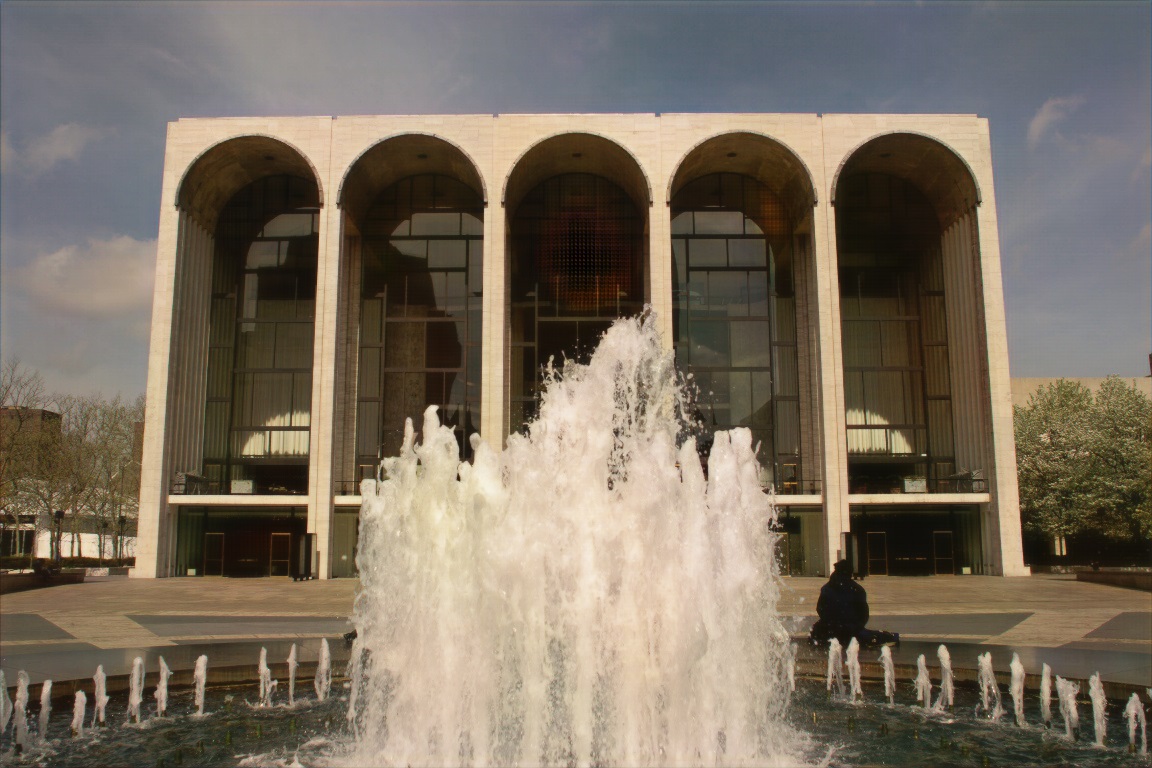}
		\caption{Using $F_{4}$}
	\end{subfigure}
	\caption{Effect of different filters.}
	\label{fig:bank}
	\vspace{-1.0em} 
\end{figure}

A significant advantage of the filter bank model is that we do not need to manually design the filters. In this section, we show some results of applying the automatically trained filters. Interestingly, each filter appears to correspond to a specific transformation. For example, the filter $F_{1}$ corresponds to brightness reduction while filter $F_{2}$ corresponds to brightness increase. This is consistent with \cite{stylebank}, except that we do not explicitly specify each filter's function.  
Figure~\ref{fig:bank} shows images generated by different filters. 

\subsection{Observations on Learned Transformation}
\label{transformation_anaylysis}

\begin{figure}[ht!]
	
	\begin{subfigure}{0.23\linewidth}
		\centering
		\includegraphics[width=0.8\linewidth]{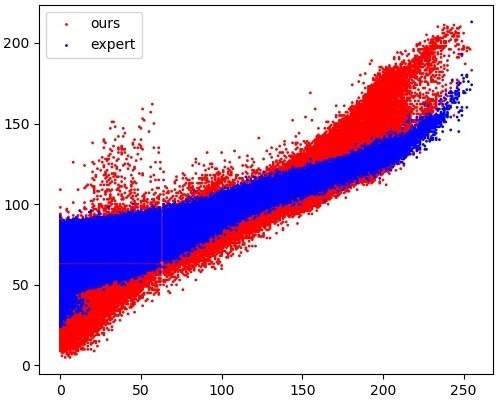}
	\end{subfigure}
	\begin{subfigure}{.23\linewidth}
		\centering
		\includegraphics[width=0.9\linewidth]{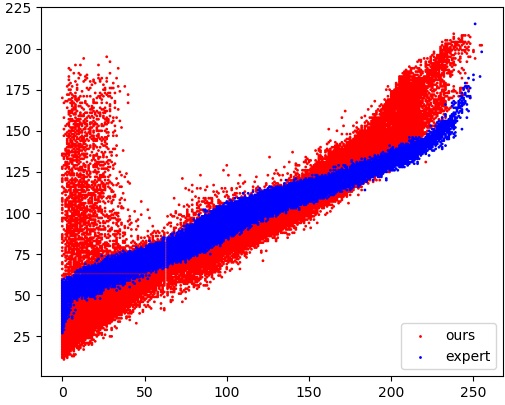}
	\end{subfigure}
	\begin{subfigure}{.23\linewidth}
		\centering
		\includegraphics[width=0.9\linewidth]{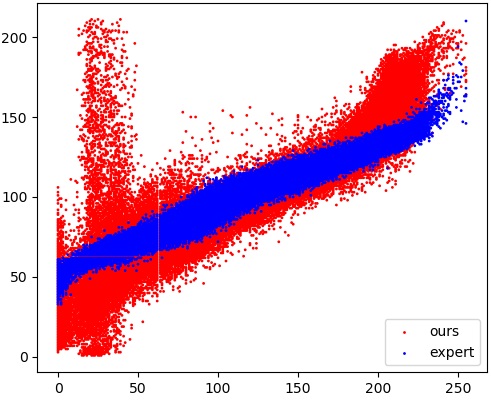}
	\end{subfigure}
	\begin{subfigure}{.23\linewidth}
		\centering
		\includegraphics[width=1.0\linewidth]{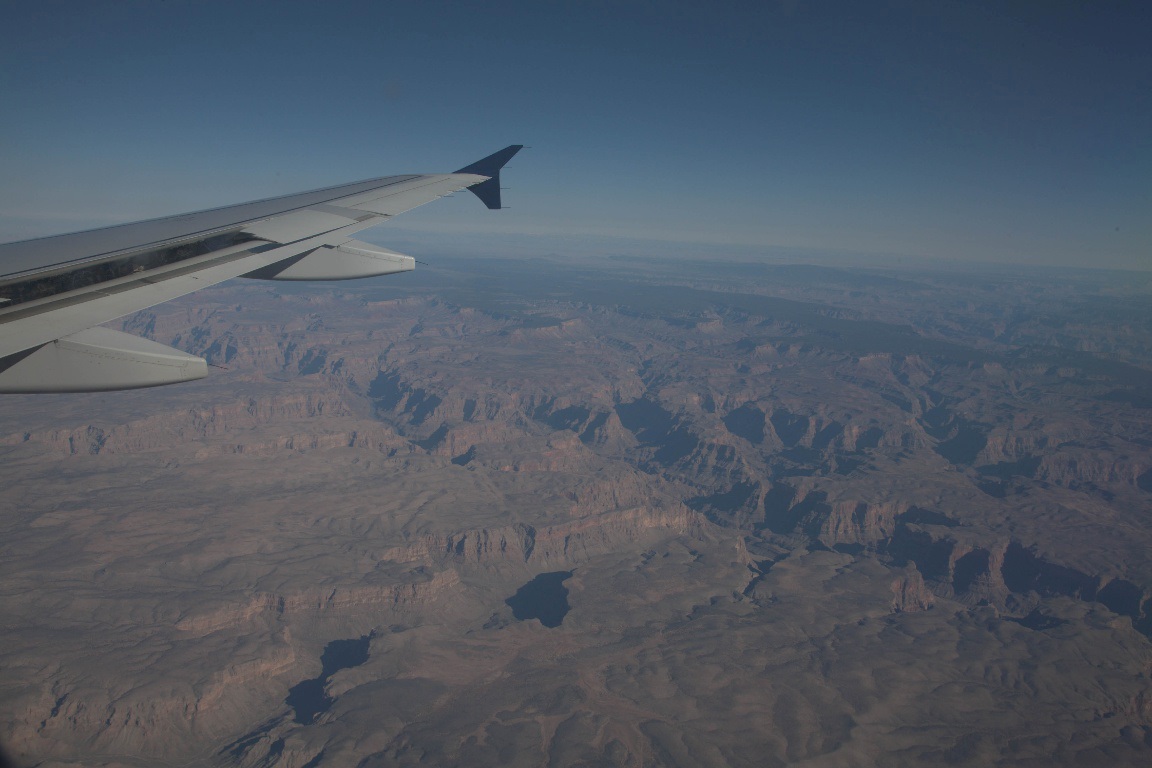}
	\end{subfigure}
	
		\begin{subfigure}{0.23\linewidth}
		\centering
		\includegraphics[width=0.8\linewidth]{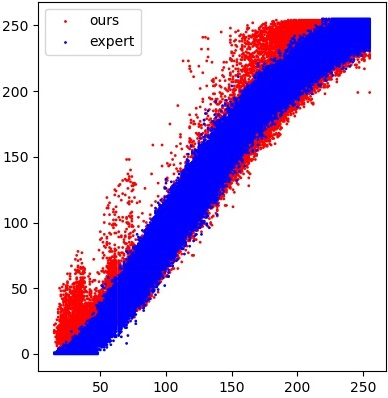}
		\caption{R channel}
	\end{subfigure}
	\begin{subfigure}{.23\linewidth}
		\centering
		\includegraphics[width=0.8\linewidth]{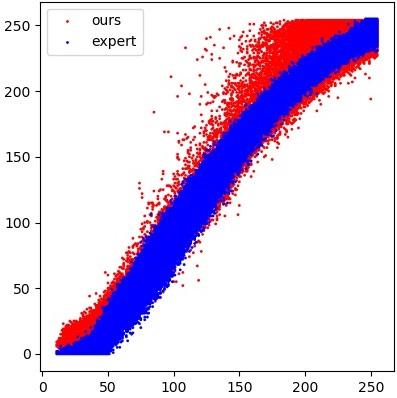}
		\caption{G channel}
	\end{subfigure}
	\begin{subfigure}{.23\linewidth}
		\centering
		\includegraphics[width=0.85\linewidth]{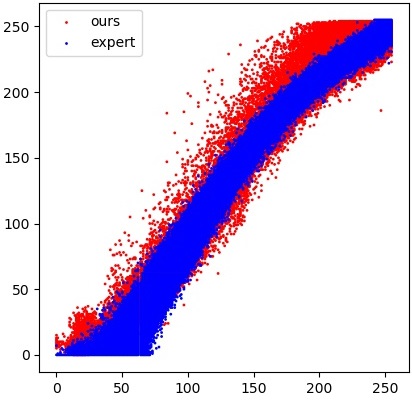}
		\caption{B channel}
	\end{subfigure}
	\begin{subfigure}{.23\linewidth}
		\centering
		\includegraphics[width=1.03\linewidth]{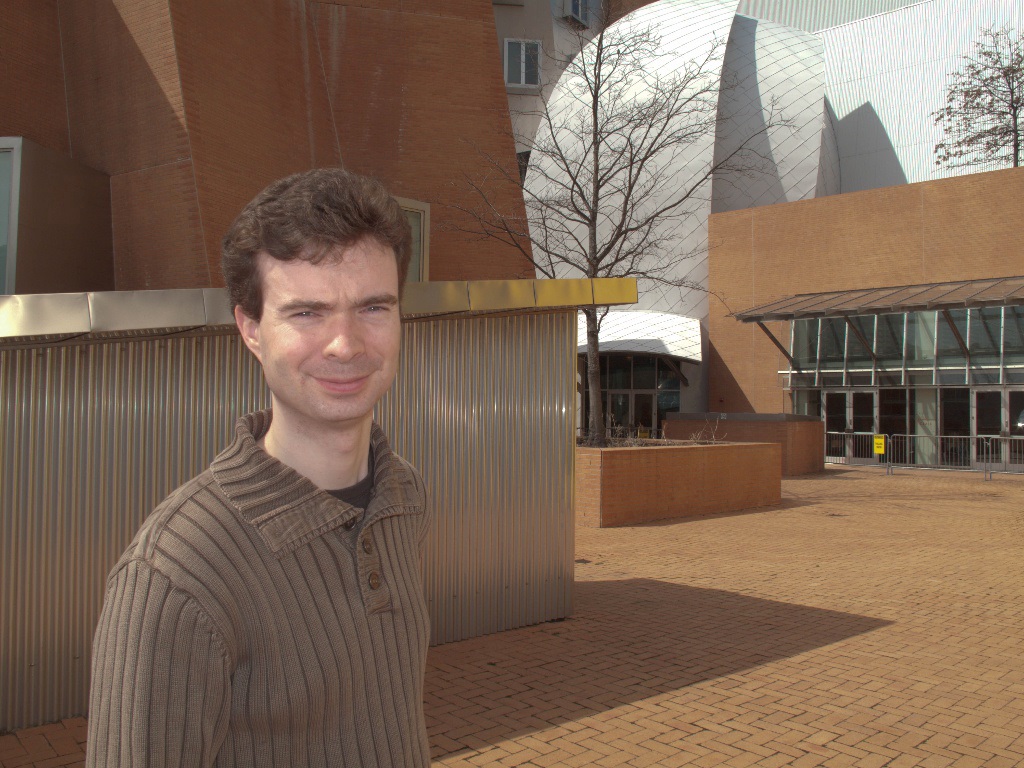}
		\caption{Original Image}
	\end{subfigure}

	\caption{RGB remapping distributions. For the expert enhanced image (expert A in MIT-Adobe 5k dataset), the mappings are almost one-to-one (in blue), while those for our edited images (in red) are not, demonstrating our editing is local.}
	\label{fig:rgb_remapping}
	\vspace{-2.0em} 
\end{figure}

Even though our models were trained on global tonal adjustments, they are able to learn local transformations. 
The RGB remapping distributions in Figure~\ref{fig:rgb_remapping} for two representative images show that our transformation, unlike its counterpart for expert A in the MIT-Adobe 5k dataset, is local. This is evident from the significantly more spread out distributions; for our method, each RGB input is mapped to wider range of outputs compared with the expert A. This results demonstrate that our models learn much more complicated mapping other than a single global mapping.       

\subsection{Using Graph RNN on Filter Bank Model}
\label{graph_rnn}

The filter bank model is a good compromise between manual effort and performance. To further investigate the influence of text encoding, we also replaced RNN with Graph RNN, more specifically, Graph GRU (Gated Recurrent Unit). The graph structure is obtained from \cite{manning2014stanford}; the last hidden state of graph RNN in both directions are used to represent the text. Please note that that conventional RNN is still used in the discriminator. Table~\ref{table:score_rnns} shows that Graph GRU performed better than plain GRU. One explanation is Graph GRU can utilize the dependency structures between different tokens and ignore the less important words in textual instructions. We also found that the Graph GRU is more effective when the text description is long and ambiguous.
For more analysis with Graph RNN, see the supplementary file.            
\begin{table}[H]
\caption{Performance comparison between different RNNs in our filter bank model.}
	\begin{center}
		\begin{tabular}{|c|c|c|}
			\hline
			& Graph GRU & GRU \\ \hline
			Standalone rating & 3.35 $\pm$ 1.24 & 3.31 $\pm$ 1.22    \\ \hline		
			Pairwise rating & 0.52 & 0.48    \\ \hline
		\end{tabular}
	\end{center}
	\label{table:score_rnns}
\end{table}

\section{Concluding Remarks}
\label{conclusion}

We show how we train a system to globally edit an image given a general textual command. To this end, we propose three models (bucket, filter bank, and end-to-end), which have different requirements in terms of initialization, memory requirements, and amount of training. Given the lack of database on image pair with text descriptions, we collected one on our own. Experimental results validate our models, and we believe our work is the first to address the general computational photography application of editing images purely through textual description. 

One current limitation is we handle only editing based on global transformations (even the learned the transformation is local). To allow object-based editing, we would need to integrate object segmentation with a natural language module \cite{nlp4seg,hong2018inferring} to our system or design a joint model which can simultaneously segment and transform~\cite{hong2018inferring}. At the same time, we found it's difficult to obtain large scale, while diverse enough data, one possible way to alleviate this issue is data augmentation~\cite{dong2017i2t2i}. Given the facts user usually prefer a series of simple, consecutive and coherent textual description, another interesting direction is to extend our work in chat environment~\cite{Das1,sharma2018chatpainter}. Finally, we can also investigate the multi-DSSM~\cite{huang2013learning} loss in our model~\cite{xu2017attngan}. We leave the development of all these functionality for future work and we believe this will be an exciting and important research topic.  


\clearpage
\bibliographystyle{splncs}
\bibliography{egbib}

\begin{thebibliography}{10}

\bibitem{pixeltone}
Laput, G., Dontcheva, M., Wilensky, G., Chang, W., Agarwala, A., Linder, J.,
  Adar, E.:
\newblock Pixeltone: A multimodal interface for image editing.
\newblock Human-Computer Interaction International Conference (HCI) (2013)

\bibitem{imgsp}
Cheng, M.M., Zheng, S., Lin, W.Y., Vineet, V., Sturgess, P., Croo, N., Mitra,
  N., Torr, P.:
\newblock Imagespirit: Verbal guided image parsing.
\newblock ACM Transactions on Graphics (2014)

\bibitem{GAN}
Goodfellow, I.J., Pouget-Abadie, J., Mirza, M., Xu, B., Warde-Farley, D.,
  Ozair, S., Courville, A., Bengio, Y.:
\newblock Generative adversarial networks.
\newblock Neural Information Processing Systems (NIPS) (2014)

\bibitem{mrf}
Kindermann, R., Snell, J.L.:
\newblock Markov random fields and their applications.
\newblock American Mathematical Society (1980)

\bibitem{chen2017language}
Chen, J., Shen, Y., Gao, J., Liu, J., Liu, X.:
\newblock Language-based image editing with recurrent attentive models.
\newblock arXiv preprint arXiv:1711.06288 (2017)

\bibitem{Seitaro2017}
Seitaro, S., Koichiro, Y., Sakriani, S., Yu, S., Satoshi, N.:
\newblock Interactive image manipulation with natural language instruction
  commands.
\newblock arXiv preprint arXiv:1802.08645 (2018)

\bibitem{nlp4seg}
Hu, R., Rohrbach, M., Darrell, T.:
\newblock Segmentation from natural language expressions.
\newblock European Conference on Computer Vision (ECCV) (2016)

\bibitem{luong2015effective}
Luong, T., Pham, H., Manning, C.D.:
\newblock Effective approaches to attention-based neural machine translation.
\newblock In: Proceedings of the 2015 Conference on Empirical Methods in
  Natural Language Processing. (2015)  1412--1421

\bibitem{peng2017cross}
Peng, N., Poon, H., Quirk, C., Toutanova, K., Yih, W.t.:
\newblock Cross-sentence n-ary relation extraction with graph lstms.
\newblock Transactions of the Association of Computational Linguistics
  \textbf{5}(1) (2017)  101--115

\bibitem{Vladimir}
Bychkovsky, V., Paris, S., Chan, E., Durand, F.:
\newblock Learning photographic global tonal adjustment with a database of
  input / output image pairs.
\newblock IEEE Conference on Computer Vision and Pattern Recognition (CVPR)
  (2011)

\bibitem{S.Hwang}
Hwang, S., Kapoor, A., Kang, S.B.:
\newblock Context-based automatic local image enhancement.
\newblock European Conference on Computer Vision (ECCV) (2012)

\bibitem{A.Kapoor}
Kapoor, A., Caicedo, J.C., Lischinski, D., Kang, S.B.:
\newblock Collaborative personalization of image enhancement.
\newblock International Journal of Computer Vision (IJCV) (2013)

\bibitem{jianzhou}
Yan, J., Lin, S., Kang, S.B., Tang, X.:
\newblock A learning-to-rank approach for image color enhancement.
\newblock IEEE Conference on Computer Vision and Pattern Recognition (CVPR)
  (2014)

\bibitem{Zhicheng}
Yan, Z., Zhang, H., Wang, B., Paris, S., Yu, Y.:
\newblock Automatic photo adjustment using deep neural networks.
\newblock ACM Transactions on Graphics (2015)

\bibitem{gatys}
Gatys, L., Ecker, A., Bethge, M.:
\newblock Image style transfer using convolutional neural networks.
\newblock IEEE Conference on Computer Vision and Pattern Recognition (CVPR)
  (2016)

\bibitem{Youngbae}
Hwang, Y., Lee, J.Y., Kweon, I.S., Kim, S.J.:
\newblock Color transfer using probabilistic moving least squares.
\newblock IEEE Conference on Computer Vision and Pattern Recognition (CVPR)
  (2014)

\bibitem{JoonYoungLee}
Lee, J.Y., Sunkavalli, K., Lin, Z., Shen, X., Kweon, I.S.:
\newblock Automatic content-aware color and tone stylization.
\newblock IEEE Conference on Computer Vision and Pattern Recognition (CVPR)
  (2016)

\bibitem{Yiming}
Liu, Y., Cohen, M., Uyttendaele, M., Rusinkiewicz, S.:
\newblock Autostyle: Automatic style transfer from image collections to
  users’ images.
\newblock Eurographics (2014)

\bibitem{st}
Vinyals, O., Toshev, A., Bengio, S., Erhan, D.:
\newblock Show and tell: A neural image caption generator.
\newblock IEEE Conference on Computer Vision and Pattern Recognition (CVPR)
  (2015)

\bibitem{minde}
Chen, X., Zitnick, C.L.:
\newblock Mind’s eye: A recurrent visual representation for image caption
  generation.
\newblock IEEE Conference on Computer Vision and Pattern Recognition (CVPR)
  (2015)

\bibitem{jeff}
Donahue, J., Darrell, T.:
\newblock Long-term recurrent convolutional networks for visual recognition and
  description.
\newblock IEEE Conference on Computer Vision and Pattern Recognition (CVPR)
  (2015)

\bibitem{chenxi}
Liu, C., Mao, J., Sha, F., Yuille, A.:
\newblock Attention correctness in neural image captioning.
\newblock IEEE Conference on Computer Vision and Pattern Recognition (CVPR)
  (2016)

\bibitem{bodai}
Dai, B., Lin, D., Urtasun, R., Fidler, S.:
\newblock Towards diverse and natural image descriptions via a conditional gan.
\newblock IEEE Conference on Computer Vision and Pattern Recognition (CVPR)
  (2017)

\bibitem{Ting}
Huang, T.H.K., Parikh, D., Vanderwende, L., Galley, M., Mitchell, M.:
\newblock Visual storytelling.
\newblock IEEE Conference on Computer Vision and Pattern Recognition (CVPR)
  (2016)

\bibitem{Venugopalan}
Venugopalan, S., Xu, H., Saenko, K.:
\newblock Translating videos to natural language using deep recurrent neural
  networks.
\newblock International Conference on Computer Vision (ICCV) (2015)

\bibitem{antol}
Antol, S., Agrawal, A., Batra, D., Zitnick, C.L., Parikh, D.:
\newblock Vqa: Visual question answering.
\newblock International Conference on Computer Vision (ICCV) (2015)

\bibitem{yezhou}
Yang, Y., Li, Y., Fermuller, C., Aloimonos, Y.:
\newblock Neural self talk: Image understanding via continuous questioning and
  answering.
\newblock Arxiv (2015)

\bibitem{Nasrin}
Mostafzadeh, N., Misra, I., Devlin, J., Mitchell, M., He, X., Vanderwende, L.:
\newblock Generating natural questions about an image.
\newblock Annual Meeting of the Association for Computational Linguistics (ACL)
  (2016)

\bibitem{Das1}
Das, A., Kottur, S., Gupta, K., Singh, A., Yadav, D., Moura, J.M.F., Parikh,
  D., Batra, D.:
\newblock Visual dialog.
\newblock IEEE Conference on Computer Vision and Pattern Recognition (CVPR)
  (2017)

\bibitem{nlp4image}
Socher, R., Karpathy, A., Le, Q.V., Manning, C.D., Ng, A.Y.:
\newblock Grounded compositional semantics for finding and describing images
  with sentences.
\newblock Transactions of the Association for Computational Linguistics (TACL)
  (2013)

\bibitem{nlor}
Hu, R., Xu, H., Rohrbach, M., Feng, J., Saenko, K., Darrell, T.:
\newblock Natural language object retrieval.
\newblock IEEE Conference on Computer Vision and Pattern Recognition (CVPR)
  (2016)

\bibitem{gtp}
Rohrbach, A.:
\newblock Grounding of textual phrases in images by reconstruction.
\newblock European Conference on Computer Vision (ECCV) (2016)

\bibitem{GATIS}
Reed, S., Akata, Z., Yan, X., Logeswaran, L., Schiele, B., Lee, H.:
\newblock Generative adversarial text-to-image synthesis.
\newblock International Conference on Machine Learning (ICML) (2016)

\bibitem{att2img}
Yan, X., Yang, J., Sohn, K., Lee, H.:
\newblock Attribute2image: Conditional image generation from visual attributes.
\newblock European Conference on Computer Vision (ECCV) (2016)

\bibitem{xu2017attngan}
Xu, T., Zhang, P., Huang, Q., Zhang, H., Gan, Z., Huang, X., He, X.:
\newblock Attngan: Fine-grained text to image generation with attentional
  generative adversarial networks.
\newblock arXiv preprint arXiv:1711.10485 (2017)

\bibitem{hong2018inferring}
Hong, S., Yang, D., Choi, J., Lee, H.:
\newblock Inferring semantic layout for hierarchical text-to-image synthesis.
\newblock arXiv preprint arXiv:1801.05091 (2018)

\bibitem{gcuod}
Mao, J.:
\newblock Generation and comprehension of unambiguous object descriptions.
\newblock IEEE Conference on Computer Vision and Pattern Recognition (CVPR)
  (2016)

\bibitem{mohit2017}
Yu, L., Tan, H., Bansal, M., Berg, T.L.:
\newblock A joint speaker-listener-reinforcer model for referring expressions.
\newblock IEEE Conference on Computer Vision and Pattern Recognition (CVPR)
  (2017)

\bibitem{cvpr2017}
Luo, R., Shakhnarovich, G.:
\newblock Comprehension-guided referring expressions.
\newblock IEEE Conference on Computer Vision and Pattern Recognition (CVPR)
  (2017)

\bibitem{inlpali}
Karpathy, A., Joulin, A., Li, F.F.:
\newblock Deep fragment embeddings for bidirectional image sentence mapping.
\newblock Conference on Neural Information Processing Systems (NIPS) (2014)

\bibitem{dvsa}
Karpathy, A., Li, F.F.:
\newblock Deep visual-semantic alignments for generating image descriptions.
\newblock IEEE Conference on Computer Vision and Pattern Recognition (CVPR)
  (2015)

\bibitem{ttico}
Kong, C., Lin, D., Bansal, M., Urtasun, R., Fidler, S.:
\newblock What are you talking about? text-to-image coreference.
\newblock IEEE Conference on Computer Vision and Pattern Recognition (CVPR)
  (2014)

\bibitem{tai2015improved}
Tai, K.S., Socher, R., Manning, C.D.:
\newblock Improved semantic representations from tree-structured long
  short-term memory networks.
\newblock In: Proceedings of the 53rd Annual Meeting of the Association for
  Computational Linguistics and the 7th International Joint Conference on
  Natural Language Processing (Volume 1: Long Papers). Volume~1. (2015)
  1556--1566

\bibitem{manning2014stanford}
Manning, C., Surdeanu, M., Bauer, J., Finkel, J., Bethard, S., McClosky, D.:
\newblock The stanford corenlp natural language processing toolkit.
\newblock In: Proceedings of 52nd annual meeting of the association for
  computational linguistics: system demonstrations. (2014)  55--60

\bibitem{c_gan}
Mirza, M., Osindero, S.:
\newblock Conditional generative adversarial nets.
\newblock Arxiv (2014)

\bibitem{instancenorm}
Ulyanov, D., Vedaldi, A., Lempitsky, V.:
\newblock Instance normalization: The missing ingredient for fast stylization.
\newblock Arxiv (2016)

\bibitem{Relu}
Nair, V., Hinton, G.E.:
\newblock Rectified linear units improve restricted boltzmann machines.
\newblock IEEE Conference on Machine Learning (ICML) (2010)

\bibitem{Perceptual_feifei}
Johnson, J., Alahi, A., Fei-Fei, L.:
\newblock Perceptual losses for real-time style transfer and super-resolution.
\newblock European Conference on Computer Vision (ECCV) (2016)

\bibitem{Photo_Realistic}
Ledig, C., Theis, L., Huszar, F., Caballero, J., Cunningham, A., Acosta, A.,
  Aitken, A., Tejani, A., Totz, J., Wang, Z., Shi, W.:
\newblock Photo-realistic single image super-resolution using a generative
  adversarial network.
\newblock IEEE Conference on Computer Vision and Pattern Recognition (CVPR)
  (2017)

\bibitem{vgg}
Simonyan, K., Zisserman, A.:
\newblock Very deep convolutional networks for large-scale image recognition.
\newblock International Conference on Learning Representations (ICLR) (2014)

\bibitem{batchnorm}
Ioffe, S., Szegedy, C.:
\newblock Batch normalization: Accelerating deep network training by reducing
  internal covariate shift.
\newblock International Conference on Machine Learning (ICML) (2015)

\bibitem{resnet}
He, K., Zhang, X., Ren, S., Sun, J.:
\newblock Deep residual learning for image recognition.
\newblock IEEE Conference on Computer Vision and Pattern Recognition (CVPR)
  (2016)

\bibitem{stylebank}
Chen, D., Yuan, L., Liao, J., Yu, N., Hua, G.:
\newblock Stylebank: An explicit representation for neural image style
  transfer.
\newblock IEEE Conference on Computer Vision and Pattern Recognition (CVPR)
  (2017)

\bibitem{ms_coco}
Lin, T.Y., Maire, M., Belongie, S., Bourdev, L., Girshick, R., Hays, J.,
  Perona, P., Ramanan, D., Zitnick, C.L., Dollár, P.:
\newblock Microsoft coco: Common objects in context.
\newblock European Conference on Computer Vision (ECCV) (2014)

\bibitem{referit}
Kazemzadeh, S., Ordonez, V., Matten, M., Berg, T.L.:
\newblock Referit game: Referring to objects in photographs of natural scenes.
\newblock Conference on Empirical Methods in Natural Language Processing
  (EMNLP) (2014)

\bibitem{ijcv2016}
Plummer, B.A.:
\newblock Flickr30k entities: Collecting region-to-phrase correspondences for
  richer image-to-sentence models.
\newblock International Journal of Computer Vision (IJCV) (2016)

\bibitem{pix2pix}
Isola, P., Zhu, J.Y., Zhou, T., Efros, A.A.:
\newblock Image-to-image translation with conditional adversarial networks.
\newblock IEEE Conference on Computer Vision and Pattern Recognition (CVPR)
  (2017)

\bibitem{Adam}
Kingma, D.P., Ba, J.:
\newblock Adam: A method for stochastic optimization.
\newblock International Conference on Learning Representations (ICLR) (2014)

\bibitem{gru}
Chung, J., Gulcehre, C., Cho, K.H., Bengio, Y.:
\newblock Empirical evaluation of gated recurrent neural networks on sequence
  modeling.
\newblock NIPS 2014 Workshop on Deep Learning (2014)

\bibitem{glove}
Pennington, J., Socher, R., Manning, C.D.:
\newblock Glove: Global vectors for word representation.
\newblock Empirical Methods in Natural Language Processing (EMNLP) (2014)

\bibitem{dong2017i2t2i}
Dong, H., Zhang, J., McIlwraith, D., Guo, Y.:
\newblock I2t2i: Learning text to image synthesis with textual data
  augmentation.
\newblock arXiv preprint arXiv:1703.06676 (2017)

\bibitem{sharma2018chatpainter}
Sharma, S., Suhubdy, D., Michalski, V., Kahou, S.E., Bengio, Y.:
\newblock Chatpainter: Improving text to image generation using dialogue.
\newblock arXiv preprint arXiv:1802.08216 (2018)

\bibitem{huang2013learning}
Huang, P.S., He, X., Gao, J., Deng, L., Acero, A., Heck, L.:
\newblock Learning deep structured semantic models for web search using
  clickthrough data.
\newblock In: Proceedings of the 22nd ACM international conference on
  Conference on information \& knowledge management, ACM (2013)  2333--2338

\end{thebibliography}

\end{document}



\pagestyle{headings}
\mainmatter

\title{Learning to Globally Edit Images\\ 
with Textual Description\\



\maketitle

In this supplementary file, we provide more information on the interface used for our data collection (Section 1) and on our implementation (Section 2). We also show additional results for our three models (Section 3).

\section{Interface for Data Collection}
\label{interface}

The interface used for our data collection (through Amazon Mechanical Turk) consists of the following parts:
\begin{itemize}
	\item Introduction page that lists instructions (Figure~\ref{fig:introduction}),
	\item Examples as guidance (Figure~\ref{fig:examples}),
	\item Qualification test to ensure the subject has some understanding of image concepts (Figure~\ref{fig:quals}), and
	\item Data collection (Figures~\ref{fig:interface}).
\end{itemize}

\begin{figure}[ht!]
	\centering
	\includegraphics[width=0.5\textwidth]{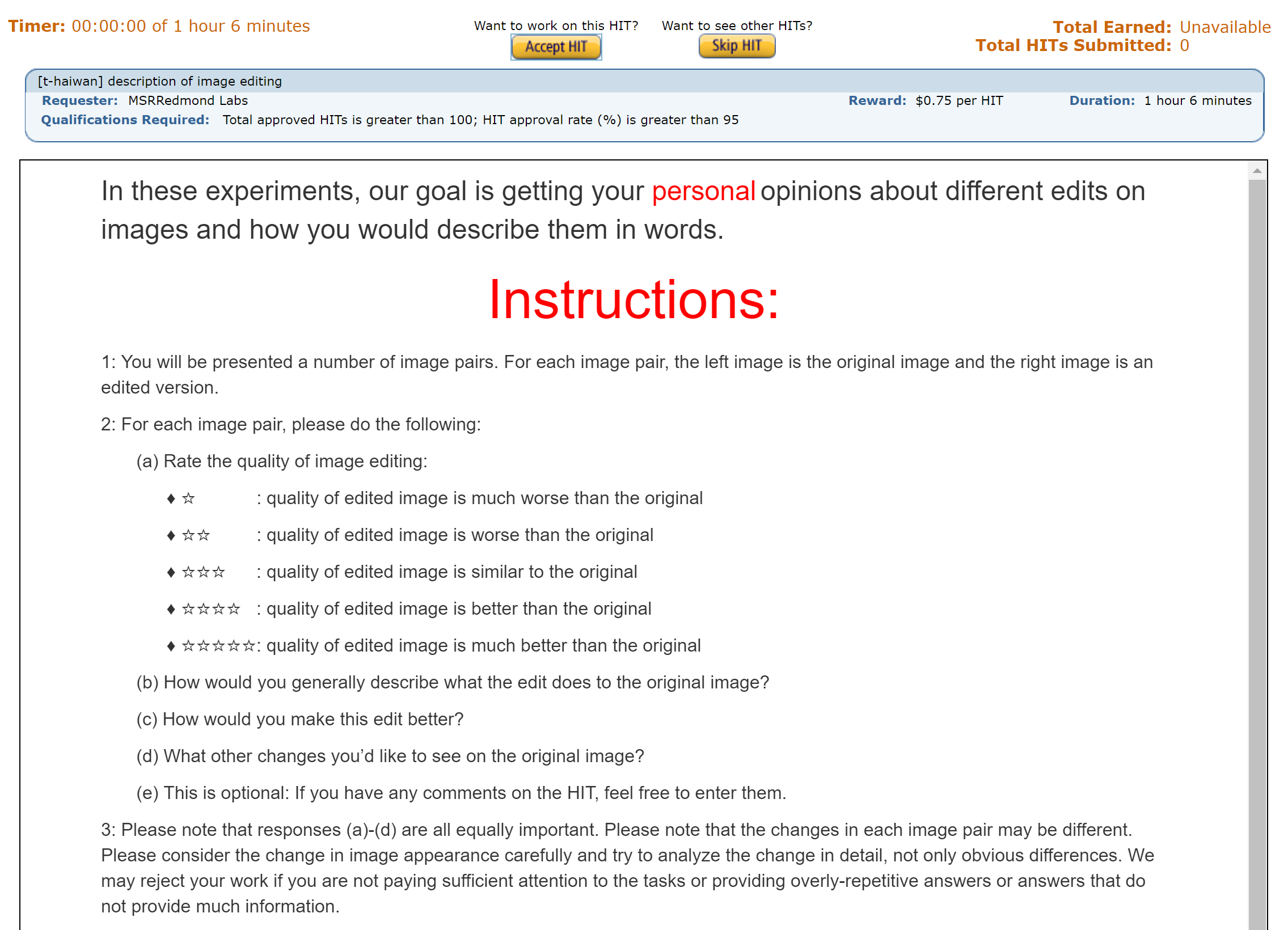}
	\caption{Instructions for data collection.}
	\label{fig:introduction}
\end{figure}

\begin{figure}[ht!]
	\centering
	\includegraphics[width=0.5\textwidth]{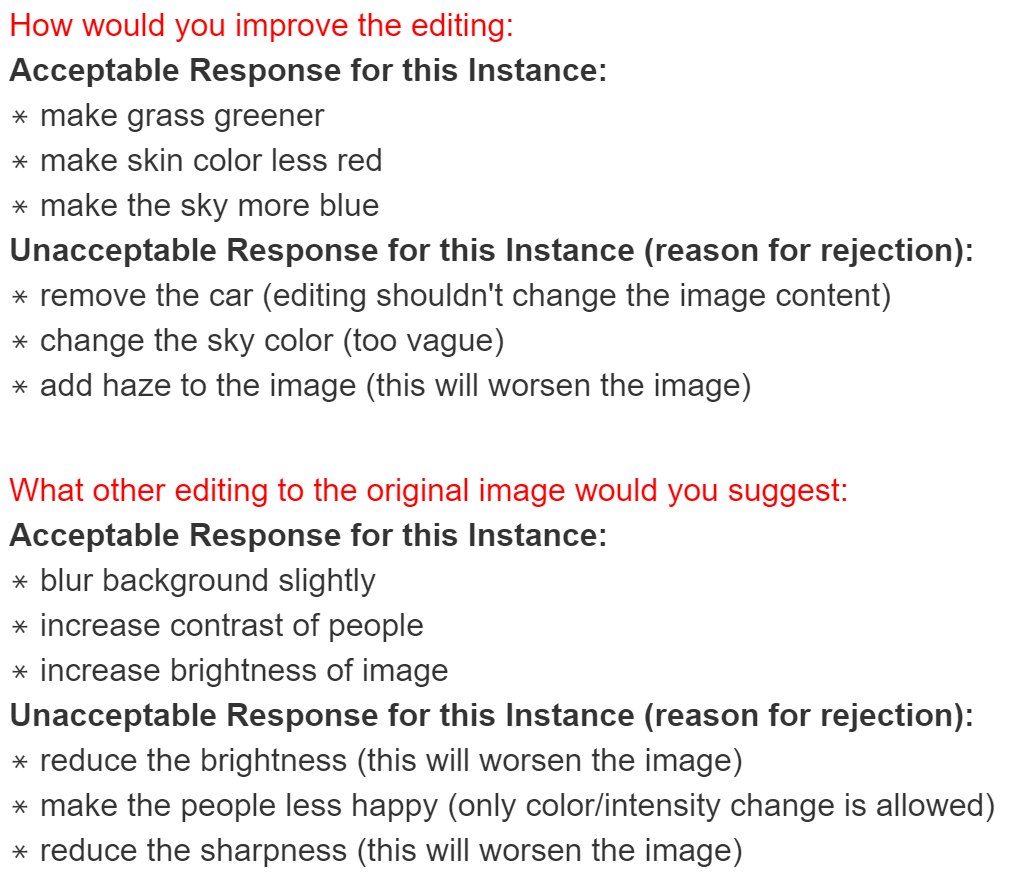}
	\caption{Examples provided as guidelines.}
	\label{fig:examples}
\end{figure}     

\begin{figure}[ht!]
	\centering
	\includegraphics[width=0.5\textwidth]{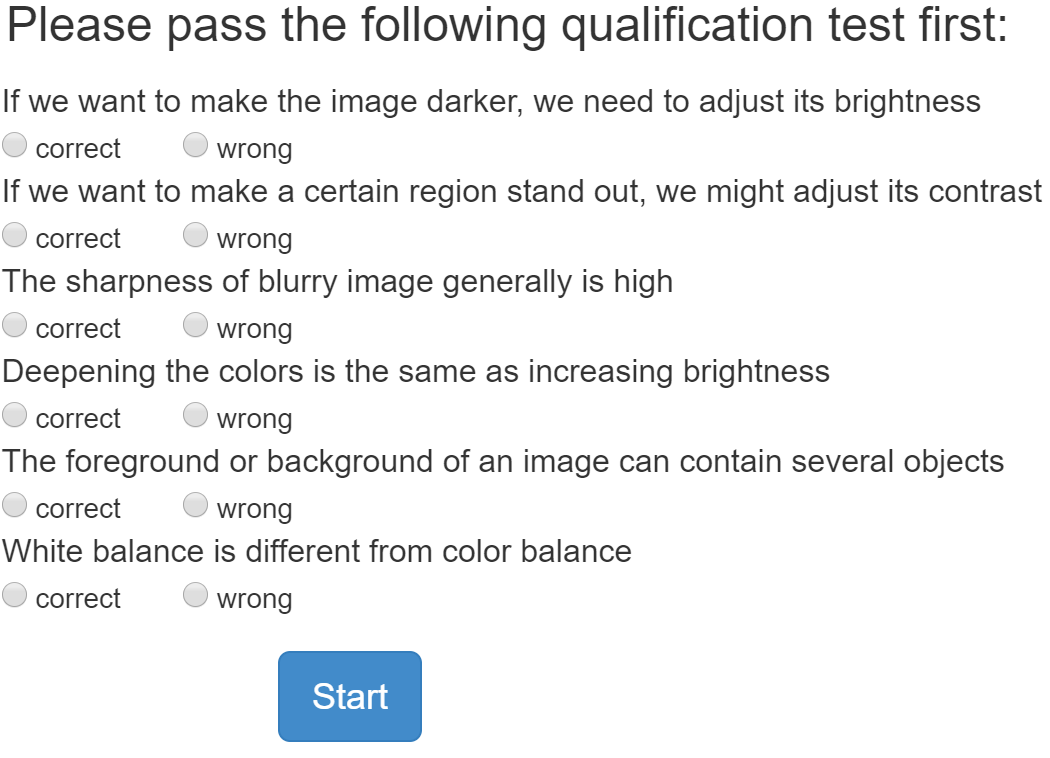}
	\caption{Qualification test.}	
	\label{fig:quals}
\end{figure}     

\begin{figure}[ht!]
	\centering
	\includegraphics[width=0.5\textwidth]{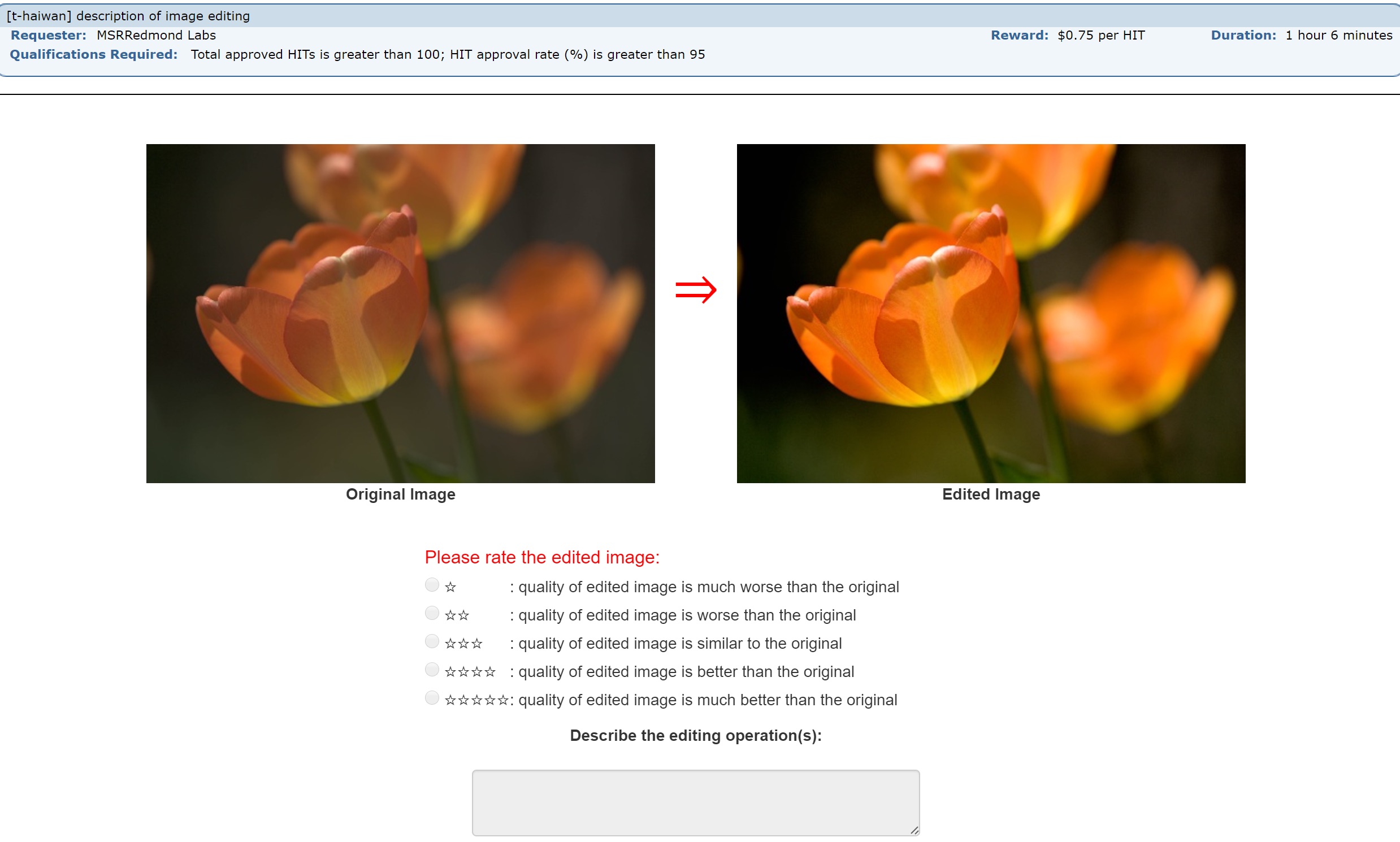}
	\caption{Interface for rating the quality of the edited image relative to the original and for providing text description.}
	\label{fig:interface}
\end{figure}

The interface used for evaluation consists of:
\begin{itemize}
	\item Instructions (Figure~\ref{fig:testmturk_intro}), 
	\item Examples as guidance (Figure~\ref{fig:testmturk_examples}), and 
	\item Rating (Figure~\ref{fig:testmturk}).
\end{itemize}

\begin{figure}[ht!]
	\centering
	\includegraphics[width=0.5\textwidth]{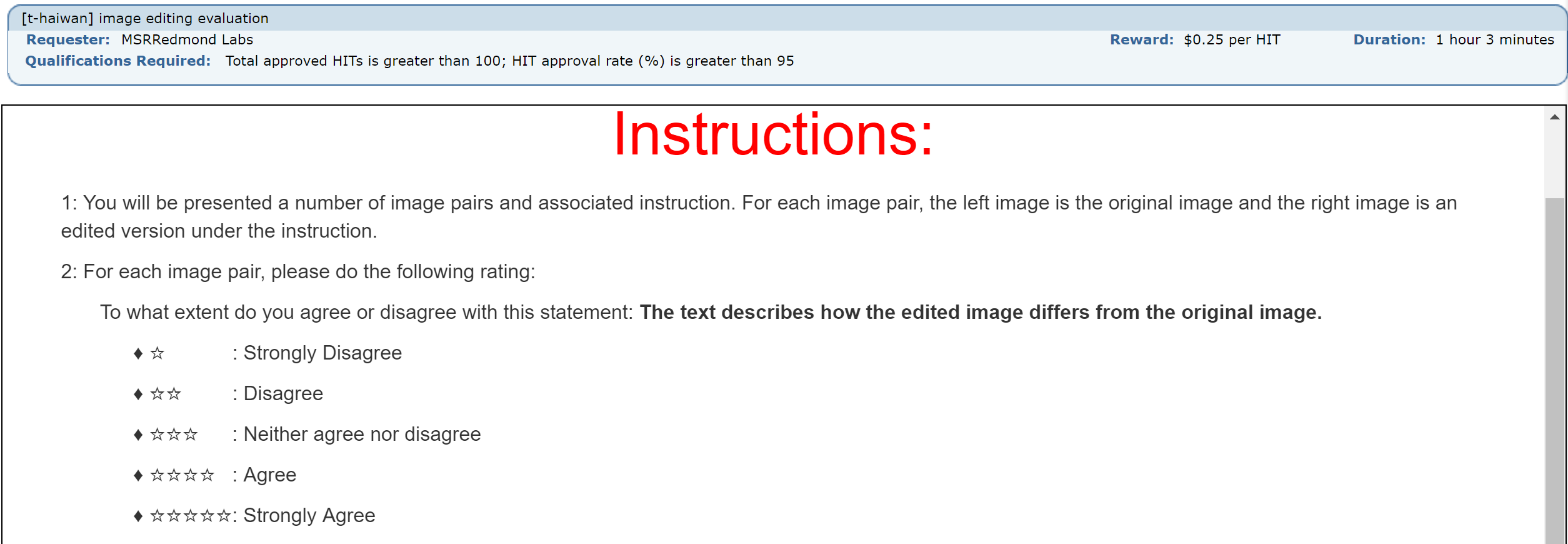}
	\caption{Introduction for rating the image editing under the text description.}
	\label{fig:testmturk_intro}	
\end{figure}     

\begin{figure}[ht!]
	\centering
	\includegraphics[width=0.5\textwidth]{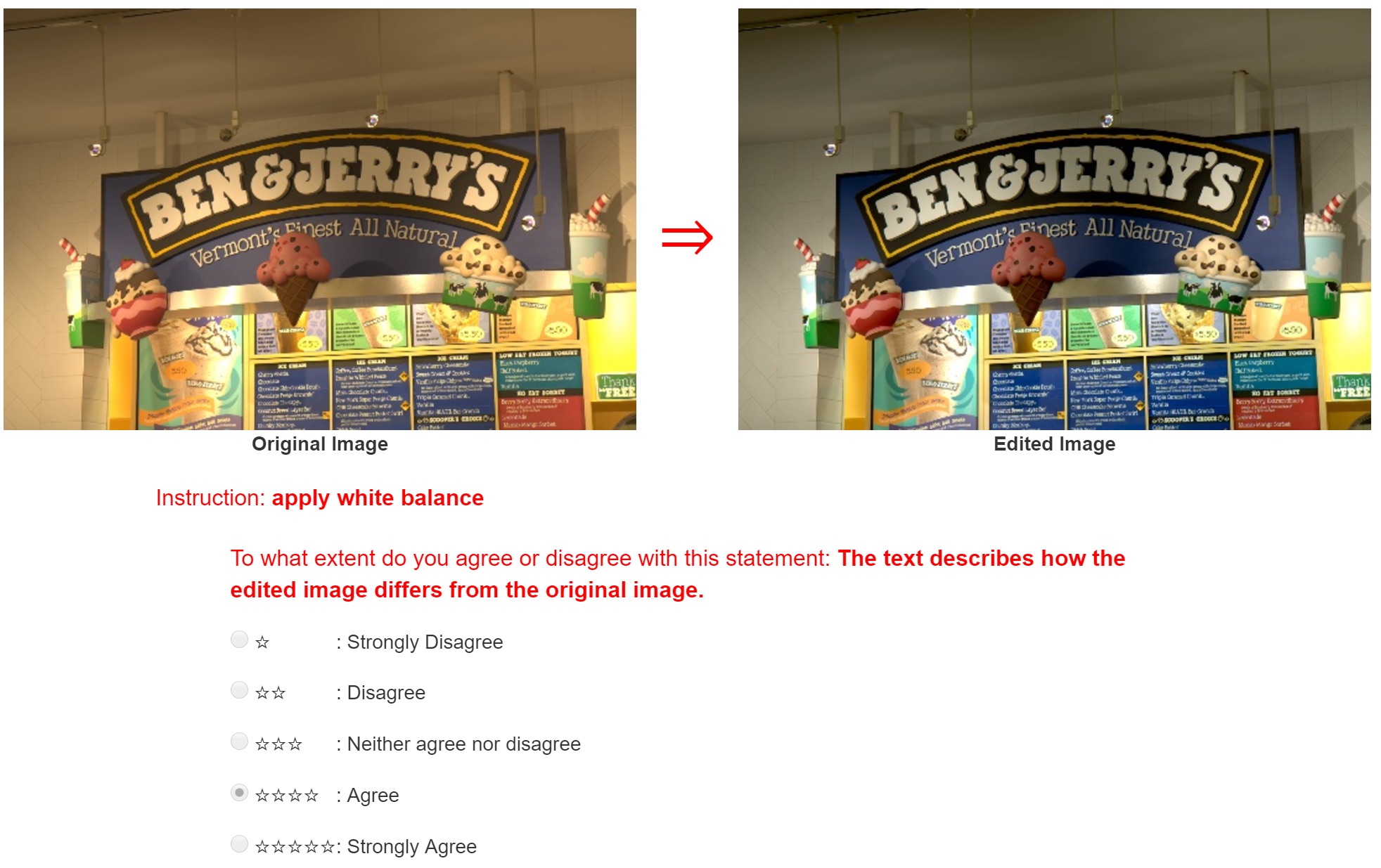}
	\caption{Examples provided for rating the image editing under the text description.}
	\label{fig:testmturk_examples}	
\end{figure}     

\begin{figure}[ht!]
	\centering
	\includegraphics[width=0.5\textwidth]{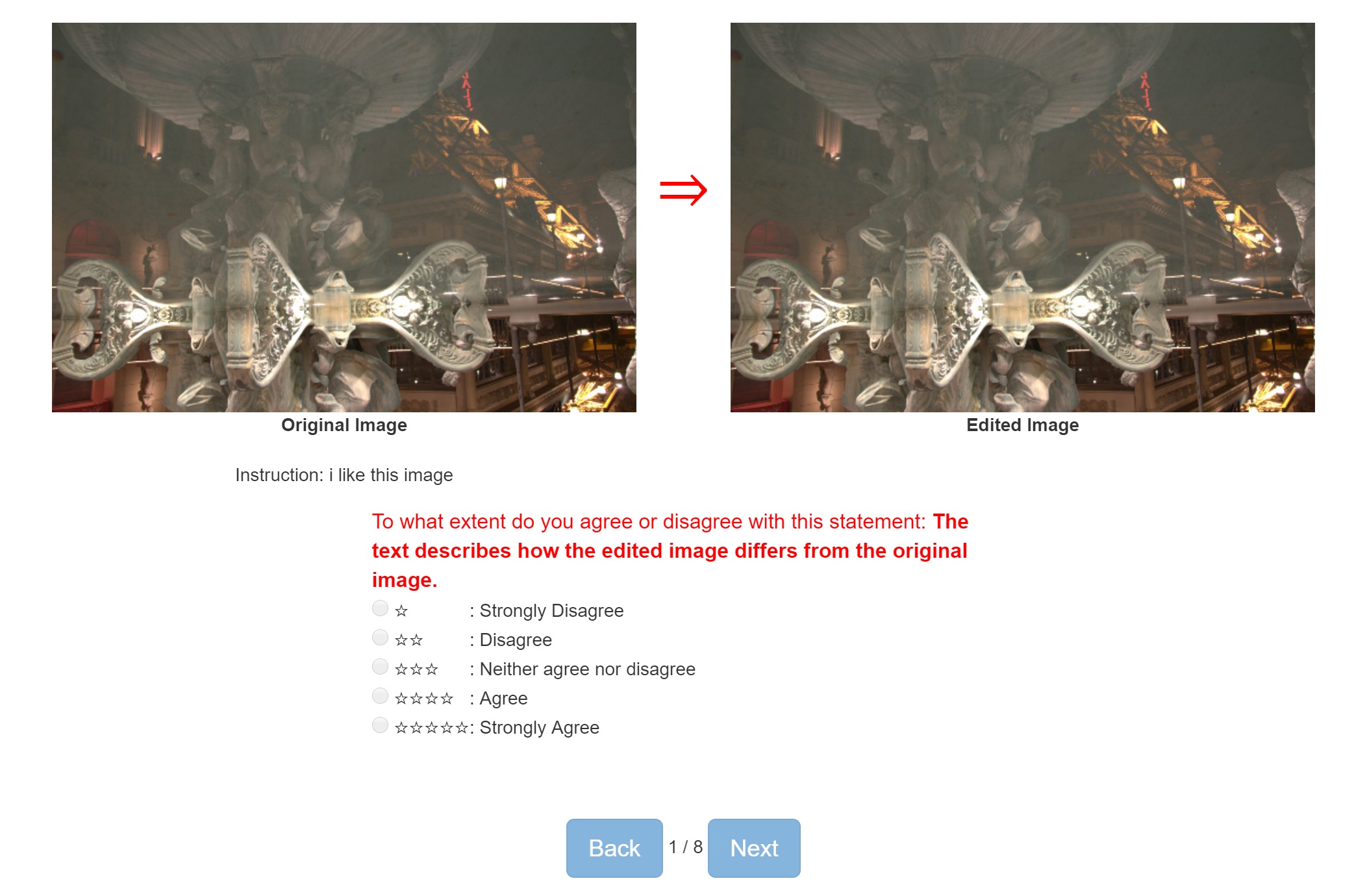}
	\caption{Interface for rating the image editing under the text description}
	\label{fig:testmturk}	
\end{figure}     

We analyzed the data collected through unigram statistics (Figure~\ref{fig:unigram}) and bigram statistics (Figure~\ref{fig:bigram}). Please note that the orange curve represents cumulative frequency.

\begin{figure}[ht!]
	\centering
	\includegraphics[width=0.5\textwidth]{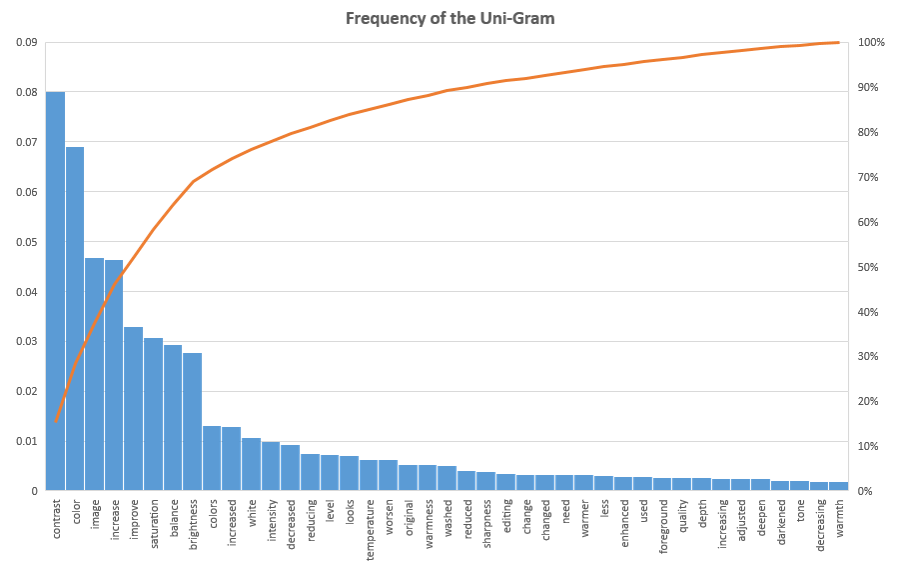}
	\caption{Unigram statistics of descriptions.}
	\label{fig:unigram}	
\end{figure}     

\begin{figure}[ht!]
	\centering
	\includegraphics[width=0.5\textwidth]{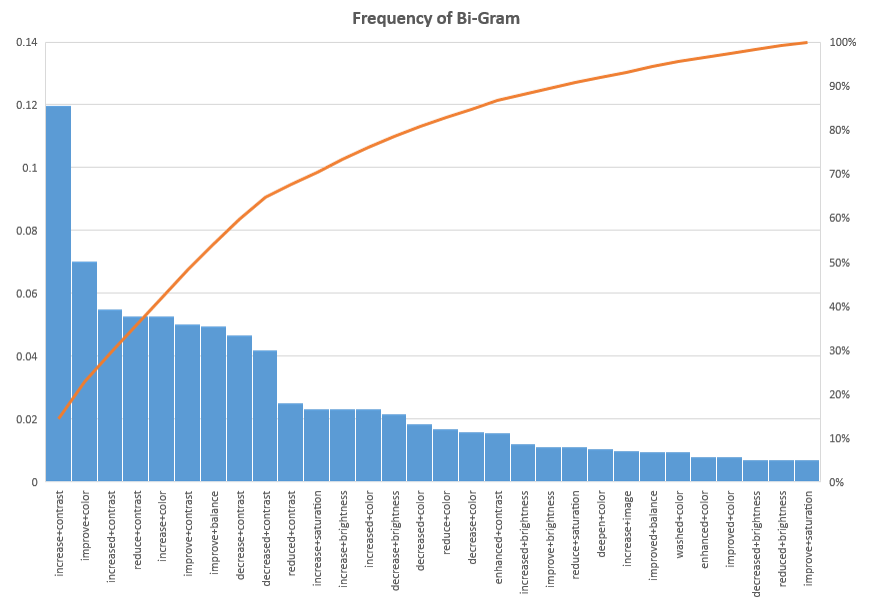}
	\caption{Bigram statistics of descriptions.}
	\label{fig:bigram}	
\end{figure}

\section{Implementation Details}
\label{exp:details}

The convolution kernels used are $4 \times 4$ spatial filters with stride 2. The downsample ratio is 2 for both encoder and discriminator, while the upsample ratio is 2 for the decoder. We find the skip connection helps to speed up the training and improve the editing performance. With the skip connections used, we have 8 Convolution-InstanceNorm-Leaky-ReLU layers. The number of filters in each layer in encoder are 64-128-256-512-512-512-512-512, while in decoder, they are 512-1024-1024-1024-1024-512-256-128. The filter numbers in discriminator are 64-128-256-512. We use the default Leaky-ReLU function without any parameter modification. 

For the RNN, we use a one-layer bi-directional GRU. All the models are trained 100 epochs, with each epoch taking around 20 minutes on a single TitanX GPU.   

We have tried a number of designs for the buckets: 
\begin{itemize}
	\item Surface form bag-of-word cluster over the descriptions,
	\item Cluster over the descriptions use sentence embedding obtained from various ways:
	\begin{itemize}
		\item Averaged on all tokens' embedding, 
		\item Element-wise product on all tokens' embedding, and
		\item Element-wise max on all tokens' embedding, 
	\end{itemize}
	\item Manual design. 
\end{itemize}
We found the manually designed the bucket to produce the best results. To design the buckets, we looked at the bigram statistics (Figure~\ref{fig:bigram}) and images in the dataset to select major image attributes, such as brightness, contrast, and white balance. We then partition the images into groups, each to be trained as a bucket, using image statistics such as difference in mean gray value and mean RGB vector between the image pairs. Please note that an image may belong in more than one group/bucket.

Initially, we found that the generator is capable to handling changes in only one direction; for example, if we train a bucket to make the image brighter, then no matter what kind of image is given, the model will always try to make the image brighter, even if the image brightness is already high. To alleviate this problem, we augment the training data: we use the same image as input and ground truth to make a new pair and we manually generate some description according to some templates as following:

\begin{itemize}
	\item \underline {it} is good; \underline {the image} is good; \underline {this picture} is \underline {amazing}; \underline {this picture} \underline {looks good};
	\item I would like to \underline {share this image}; I would like to \underline {send this image to my friend};
	\item the tone in this image is \underline {good}; the tone in this image is \underline {perfect};    
\end{itemize}	
we can replace the underlined text with any paraphrases.

To sample a random text for the discriminator (Section 3.1 in the paper), we first calculate the confusion matrix between most frequent unigram (Figure~\ref{fig:unigram}) by checking their co-occurrence in the description. Given a description, we first find the most unrelated unigrams and then find a description that contains these unigrams.

\section{Additional Results}
\label{exp:result}

In this section, we show more results for generation of edited images given an input image and text description using our three models: 
bucket model, fusion version (Figure~\ref{exp:bucket}), 
bucket model, argmax version (Figure~\ref{exp:bucket:argmax}), 
filter bank model (Figure~\ref{exp:filterbank}),
and end-to-end model (Figure \ref{exp:end-to-end}).

\begin{figure*}
	\centering
	
	\begin{subfigure}{.27\linewidth}
		\centering
		\includegraphics[width=.9\linewidth]{figs/bucket_fusion/expertA-original_shotzero-a4658-Duggan_090201_4929_real_A.jpg}
		\caption{Input + ``decrease saturation"}
		\label{fig:sfig1}
	\end{subfigure}%
	\begin{subfigure}{.27\linewidth}
		\centering
		\includegraphics[width=.9\linewidth]{figs/bucket_fusion/expertA-original_shotzero-a4658-Duggan_090201_4929_fake_B.jpg}
		\caption{Output}
		\label{fig:sfig2}
	\end{subfigure}
	\begin{subfigure}{.27\linewidth}
		\centering
		\includegraphics[width=.9\linewidth]{figs/bucket_fusion/expertA-original_shotzero-a4658-Duggan_090201_4929_real_B.jpg}
		\caption{Ground Truth}
		\label{fig:sfig3}
	\end{subfigure}
	
	\begin{subfigure}{.27\linewidth}
		\centering
		\includegraphics[width=.9\linewidth]{figs/bucket_fusion/original_shotzero-expertA-a4636-Duggan_080216_5303_real_A.jpg}
		\caption{Input + ``improve color balance"}
		\label{fig:sfig1}
	\end{subfigure}%
	\begin{subfigure}{.27\linewidth}
		\centering
		\includegraphics[width=.9\linewidth]{figs/bucket_fusion/original_shotzero-expertA-a4636-Duggan_080216_5303_fake_B.jpg}
		\caption{Output}
		\label{fig:sfig2}
	\end{subfigure}
	\begin{subfigure}{.27\linewidth}
		\centering
		\includegraphics[width=.9\linewidth]{figs/bucket_fusion/original_shotzero-expertA-a4636-Duggan_080216_5303_real_B.jpg}
		\caption{Ground Truth}
		\label{fig:sfig3}
	\end{subfigure}
	
	\begin{subfigure}{.27\linewidth}
		\centering
		\includegraphics[width=.9\linewidth]{figs/bucket_fusion/original_shotzero-expertA-a4658-Duggan_090201_4929_real_A.jpg}
		\caption{Input + ``increase saturation"}
		\label{fig:sfig1}
	\end{subfigure}%
	\begin{subfigure}{.27\linewidth}
		\centering
		\includegraphics[width=.9\linewidth]{figs/bucket_fusion/original_shotzero-expertA-a4658-Duggan_090201_4929_fake_B.jpg}
		\caption{Output}
		\label{fig:sfig2}
	\end{subfigure}
	\begin{subfigure}{.27\linewidth}
		\centering
		\includegraphics[width=.9\linewidth]{figs/bucket_fusion/original_shotzero-expertA-a4658-Duggan_090201_4929_real_B.jpg}
		\caption{Ground Truth}
		\label{fig:sfig3}
	\end{subfigure}
	
	\begin{subfigure}{.27\linewidth}
		\centering
		\includegraphics[width=.9\linewidth]{figs/bucket_fusion/original_shotzero-expertA-a4775-DSC_0040_real_A.jpg}
		\caption{Input + ``color balance improved, sharpness improved"}
		\label{fig:sfig1}
	\end{subfigure}%
	\begin{subfigure}{.27\linewidth}
		\centering
		\includegraphics[width=.9\linewidth]{figs/bucket_fusion/original_shotzero-expertA-a4775-DSC_0040_fake_B.jpg}
		\caption{Output}
		\label{fig:sfig2}
	\end{subfigure}
	\begin{subfigure}{.27\linewidth}
		\centering
		\includegraphics[width=.9\linewidth]{figs/bucket_fusion/original_shotzero-expertA-a4775-DSC_0040_real_B.jpg}
		\caption{Ground Truth}
		\label{fig:sfig3}
	\end{subfigure}

	\begin{subfigure}{.27\linewidth}
		\centering
		\includegraphics[width=.9\linewidth]{figs/bucket_fusion/original_shotzero-expertA-a4864-jmacdscf0536_real_A.jpg}
		\caption{Input + ``saturation raised; colors are much deeper"}
		\label{fig:sfig1}
	\end{subfigure}%
	\begin{subfigure}{.27\linewidth}
		\centering
		\includegraphics[width=.9\linewidth]{figs/bucket_fusion/original_shotzero-expertA-a4864-jmacdscf0536_fake_B.jpg}
		\caption{Output}
		\label{fig:sfig2}
	\end{subfigure}
	\begin{subfigure}{.27\linewidth}
		\centering
		\includegraphics[width=.9\linewidth]{figs/bucket_fusion/original_shotzero-expertA-a4864-jmacdscf0536_real_B.jpg}
		\caption{Ground Truth}
		\label{fig:sfig3}
	\end{subfigure}
	
	\begin{subfigure}{.27\linewidth}
		\centering
		\includegraphics[width=.9\linewidth]{figs/bucket_fusion/original_shotzero-expertA-a4967-kme_2360_real_A.jpg}
		\caption{Input + ``color increased and skin tone of human is also good"}
		\label{fig:sfig1}
	\end{subfigure}%
	\begin{subfigure}{.27\linewidth}
		\centering
		\includegraphics[width=.9\linewidth]{figs/bucket_fusion/original_shotzero-expertA-a4967-kme_2360_fake_B.jpg}
		\caption{Output}
		\label{fig:sfig2}
	\end{subfigure}
	\begin{subfigure}{.27\linewidth}
		\centering
		\includegraphics[width=.9\linewidth]{figs/bucket_fusion/original_shotzero-expertA-a4967-kme_2360_real_B.jpg}
		\caption{Ground Truth}
		\label{fig:sfig3}
	\end{subfigure}
	
	\caption{More results for our bucket (fusion) model.}
	\label{exp:bucket}
	\vspace{-0.2em} 
\end{figure*}

\begin{figure*}
	\centering
	
	\begin{subfigure}{.27\linewidth}
		\centering
		\includegraphics[width=.9\linewidth]{figs/bucket_argmax/expertA-original_shotzero-a4658-Duggan_090201_4929_real_A.jpg}
		\caption{Input + ``decrease saturation"}
		\label{fig:sfig1}
	\end{subfigure}%
	\begin{subfigure}{.27\linewidth}
		\centering
		\includegraphics[width=.9\linewidth]{figs/bucket_argmax/expertA-original_shotzero-a4658-Duggan_090201_4929_fake_B.jpg}
		\caption{Output}
		\label{fig:sfig2}
	\end{subfigure}
	\begin{subfigure}{.27\linewidth}
		\centering
		\includegraphics[width=.9\linewidth]{figs/bucket_argmax/expertA-original_shotzero-a4658-Duggan_090201_4929_real_B.jpg}
		\caption{Ground Truth}
		\label{fig:sfig3}
	\end{subfigure}
	
	\begin{subfigure}{.27\linewidth}
		\centering
		\includegraphics[width=.9\linewidth]{figs/bucket_argmax/original_shotzero-expertA-a4636-Duggan_080216_5303_real_A.jpg}
		\caption{Input + ``improve color balance"}
		\label{fig:sfig1}
	\end{subfigure}%
	\begin{subfigure}{.27\linewidth}
		\centering
		\includegraphics[width=.9\linewidth]{figs/bucket_argmax/original_shotzero-expertA-a4636-Duggan_080216_5303_fake_B.jpg}
		\caption{Output}
		\label{fig:sfig2}
	\end{subfigure}
	\begin{subfigure}{.27\linewidth}
		\centering
		\includegraphics[width=.9\linewidth]{figs/bucket_argmax/original_shotzero-expertA-a4636-Duggan_080216_5303_real_B.jpg}
		\caption{Ground Truth}
		\label{fig:sfig3}
	\end{subfigure}
	
	\begin{subfigure}{.27\linewidth}
		\centering
		\includegraphics[width=.9\linewidth]{figs/bucket_argmax/original_shotzero-expertA-a4658-Duggan_090201_4929_real_A.jpg}
		\caption{Input + ``increase saturation"}
		\label{fig:sfig1}
	\end{subfigure}%
	\begin{subfigure}{.27\linewidth}
		\centering
		\includegraphics[width=.9\linewidth]{figs/bucket_argmax/original_shotzero-expertA-a4658-Duggan_090201_4929_fake_B.jpg}
		\caption{Output}
		\label{fig:sfig2}
	\end{subfigure}
	\begin{subfigure}{.27\linewidth}
		\centering
		\includegraphics[width=.9\linewidth]{figs/bucket_argmax/original_shotzero-expertA-a4658-Duggan_090201_4929_real_B.jpg}
		\caption{Ground Truth}
		\label{fig:sfig3}
	\end{subfigure}
	
	\begin{subfigure}{.27\linewidth}
		\centering
		\includegraphics[width=.9\linewidth]{figs/bucket_argmax/original_shotzero-expertA-a4775-DSC_0040_real_A.jpg}
		\caption{Input + ``color balance improved, sharpness improved"}
		\label{fig:sfig1}
	\end{subfigure}%
	\begin{subfigure}{.27\linewidth}
		\centering
		\includegraphics[width=.9\linewidth]{figs/bucket_argmax/original_shotzero-expertA-a4775-DSC_0040_fake_B.jpg}
		\caption{Output}
		\label{fig:sfig2}
	\end{subfigure}
	\begin{subfigure}{.27\linewidth}
		\centering
		\includegraphics[width=.9\linewidth]{figs/bucket_argmax/original_shotzero-expertA-a4775-DSC_0040_real_B.jpg}
		\caption{Ground Truth}
		\label{fig:sfig3}
	\end{subfigure}

	\begin{subfigure}{.27\linewidth}
		\centering
		\includegraphics[width=.9\linewidth]{figs/bucket_argmax/original_shotzero-expertA-a4864-jmacdscf0536_real_A.jpg}
		\caption{Input + ``saturation raised; colors are much deeper"}
		\label{fig:sfig1}
	\end{subfigure}%
	\begin{subfigure}{.27\linewidth}
		\centering
		\includegraphics[width=.9\linewidth]{figs/bucket_argmax/original_shotzero-expertA-a4864-jmacdscf0536_fake_B.jpg}
		\caption{Output}
		\label{fig:sfig2}
	\end{subfigure}
	\begin{subfigure}{.27\linewidth}
		\centering
		\includegraphics[width=.9\linewidth]{figs/bucket_argmax/original_shotzero-expertA-a4864-jmacdscf0536_real_B.jpg}
		\caption{Ground Truth}
		\label{fig:sfig3}
	\end{subfigure}
	
	\begin{subfigure}{.27\linewidth}
		\centering
		\includegraphics[width=.9\linewidth]{figs/bucket_argmax/original_shotzero-expertA-a4967-kme_2360_real_A.jpg}
		\caption{Input + ``color increased and skin tone of human is also good"}
		\label{fig:sfig1}
	\end{subfigure}%
	\begin{subfigure}{.27\linewidth}
		\centering
		\includegraphics[width=.9\linewidth]{figs/bucket_argmax/original_shotzero-expertA-a4967-kme_2360_fake_B.jpg}
		\caption{Output}
		\label{fig:sfig2}
	\end{subfigure}
	\begin{subfigure}{.27\linewidth}
		\centering
		\includegraphics[width=.9\linewidth]{figs/bucket_argmax/original_shotzero-expertA-a4967-kme_2360_real_B.jpg}
		\caption{Ground Truth}
		\label{fig:sfig3}
	\end{subfigure}
	
	\caption{More results for our bucket (argmax) model.}
	\label{exp:bucket:argmax}
	\vspace{-0.2em} 
\end{figure*}

\begin{figure*}
	\centering
	\begin{subfigure}{.27\linewidth}
		\centering
		\includegraphics[width=.9\linewidth]{figs/filterbank/expertA-original_shotzero-a4658-Duggan_090201_4929_real_A}
		\caption{Input + ``decrease saturation"}
		\label{fig:sfig1}
	\end{subfigure}%
	\begin{subfigure}{.27\linewidth}
		\centering
		\includegraphics[width=.9\linewidth]{figs/filterbank/expertA-original_shotzero-a4658-Duggan_090201_4929_fake_B}
		\caption{Output}
		\label{fig:sfig2}
	\end{subfigure}
	\begin{subfigure}{.27\linewidth}
		\centering
		\includegraphics[width=.9\linewidth]{figs/filterbank/expertA-original_shotzero-a4658-Duggan_090201_4929_real_B}
		\caption{Ground Truth}
		\label{fig:sfig3}
	\end{subfigure}
	
	\begin{subfigure}{.27\linewidth}
		\centering
		\includegraphics[width=.9\linewidth]{figs/filterbank/original_shotzero-expertA-a4636-Duggan_080216_5303_real_A}
		\caption{Input + ``improve color balance"}
		\label{fig:sfig1}
	\end{subfigure}%
	\begin{subfigure}{.27\linewidth}
		\centering
		\includegraphics[width=.9\linewidth]{figs/filterbank/original_shotzero-expertA-a4636-Duggan_080216_5303_fake_B}
		\caption{Output}
		\label{fig:sfig2}
	\end{subfigure}
	\begin{subfigure}{.27\linewidth}
		\centering
		\includegraphics[width=.9\linewidth]{figs/filterbank/original_shotzero-expertA-a4636-Duggan_080216_5303_real_B}
		\caption{Ground Truth}
		\label{fig:sfig3}
	\end{subfigure}
	
	\begin{subfigure}{.27\linewidth}
		\centering
		\includegraphics[width=.9\linewidth]{figs/filterbank/original_shotzero-expertA-a4658-Duggan_090201_4929_real_A}
		\caption{Input + ``increase saturation"}
		\label{fig:sfig1}
	\end{subfigure}%
	\begin{subfigure}{.27\linewidth}
		\centering
		\includegraphics[width=.9\linewidth]{figs/filterbank/original_shotzero-expertA-a4658-Duggan_090201_4929_fake_B}
		\caption{Output}
		\label{fig:sfig2}
	\end{subfigure}
	\begin{subfigure}{.27\linewidth}
		\centering
		\includegraphics[width=.9\linewidth]{figs/filterbank/original_shotzero-expertA-a4658-Duggan_090201_4929_real_B}
		\caption{Ground Truth}
		\label{fig:sfig3}
	\end{subfigure}
	
	\begin{subfigure}{.27\linewidth}
		\centering
		\includegraphics[width=.9\linewidth]{figs/filterbank/original_shotzero-expertA-a4775-DSC_0040_real_A}
		\caption{Input + ``color balance improved, sharpness improved"}
		\label{fig:sfig1}
	\end{subfigure}%
	\begin{subfigure}{.27\linewidth}
		\centering
		\includegraphics[width=.9\linewidth]{figs/filterbank/original_shotzero-expertA-a4775-DSC_0040_fake_B}
		\caption{Output}
		\label{fig:sfig2}
	\end{subfigure}
	\begin{subfigure}{.27\linewidth}
		\centering
		\includegraphics[width=.9\linewidth]{figs/filterbank/original_shotzero-expertA-a4775-DSC_0040_real_B}
		\caption{Ground Truth}
		\label{fig:sfig3}
	\end{subfigure}

	\begin{subfigure}{.27\linewidth}
		\centering
		\includegraphics[width=.9\linewidth]{figs/filterbank/original_shotzero-expertA-a4864-jmacdscf0536_real_A}
		\caption{Input + ``saturation raised; colors are much deeper"}
		\label{fig:sfig1}
	\end{subfigure}%
	\begin{subfigure}{.27\linewidth}
		\centering
		\includegraphics[width=.9\linewidth]{figs/filterbank/original_shotzero-expertA-a4864-jmacdscf0536_fake_B}
		\caption{Output}
		\label{fig:sfig2}
	\end{subfigure}
	\begin{subfigure}{.27\linewidth}
		\centering
		\includegraphics[width=.9\linewidth]{figs/filterbank/original_shotzero-expertA-a4864-jmacdscf0536_real_B}
		\caption{Ground Truth}
		\label{fig:sfig3}
	\end{subfigure}
	
	\begin{subfigure}{.27\linewidth}
		\centering
		\includegraphics[width=.9\linewidth]{figs/filterbank/original_shotzero-expertA-a4967-kme_2360_real_A}
		\caption{Input + ``color increased and skin tone of human is also good"}
		\label{fig:sfig1}
	\end{subfigure}%
	\begin{subfigure}{.27\linewidth}
		\centering
		\includegraphics[width=.9\linewidth]{figs/filterbank/original_shotzero-expertA-a4967-kme_2360_fake_B}
		\caption{Output}
		\label{fig:sfig2}
	\end{subfigure}
	\begin{subfigure}{.27\linewidth}
		\centering
		\includegraphics[width=.9\linewidth]{figs/filterbank/original_shotzero-expertA-a4967-kme_2360_real_B}
		\caption{Ground Truth}
		\label{fig:sfig3}
	\end{subfigure}
	
	\caption{More results for our filter bank model.}
	\label{exp:filterbank}
	\vspace{-0.2em} 
\end{figure*}

\begin{figure*}
	\centering
	
	\begin{subfigure}{.27\linewidth}
		\centering
		\includegraphics[width=.9\linewidth]{figs/end-to-end/expertA-original_shotzero-a4658-Duggan_090201_4929_real_A}
		\caption{Input + ``decrease saturation"}
		\label{fig:sfig1}
	\end{subfigure}%
	\begin{subfigure}{.27\linewidth}
		\centering
		\includegraphics[width=.9\linewidth]{figs/end-to-end/expertA-original_shotzero-a4658-Duggan_090201_4929_fake_B}
		\caption{Output}
		\label{fig:sfig2}
	\end{subfigure}
	\begin{subfigure}{.27\linewidth}
		\centering
		\includegraphics[width=.9\linewidth]{figs/end-to-end/expertA-original_shotzero-a4658-Duggan_090201_4929_real_B}
		\caption{Ground Truth}
		\label{fig:sfig3}
	\end{subfigure}
	
	\begin{subfigure}{.27\linewidth}
		\centering
		\includegraphics[width=.9\linewidth]{figs/end-to-end/original_shotzero-expertA-a4636-Duggan_080216_5303_real_A}
		\caption{Input + ``improve color balance"}
		\label{fig:sfig1}
	\end{subfigure}%
	\begin{subfigure}{.27\linewidth}
		\centering
		\includegraphics[width=.9\linewidth]{figs/end-to-end/original_shotzero-expertA-a4636-Duggan_080216_5303_fake_B}
		\caption{Output}
		\label{fig:sfig2}
	\end{subfigure}
	\begin{subfigure}{.27\linewidth}
		\centering
		\includegraphics[width=.9\linewidth]{figs/end-to-end/original_shotzero-expertA-a4636-Duggan_080216_5303_real_B}
		\caption{Ground Truth}
		\label{fig:sfig3}
	\end{subfigure}
	
	\begin{subfigure}{.27\linewidth}
		\centering
		\includegraphics[width=.9\linewidth]{figs/end-to-end/original_shotzero-expertA-a4658-Duggan_090201_4929_real_A}
		\caption{Input + ``increase saturation"}
		\label{fig:sfig1}
	\end{subfigure}%
	\begin{subfigure}{.27\linewidth}
		\centering
		\includegraphics[width=.9\linewidth]{figs/end-to-end/original_shotzero-expertA-a4658-Duggan_090201_4929_fake_B}
		\caption{Output}
		\label{fig:sfig2}
	\end{subfigure}
	\begin{subfigure}{.27\linewidth}
		\centering
		\includegraphics[width=.9\linewidth]{figs/end-to-end/original_shotzero-expertA-a4658-Duggan_090201_4929_real_B}
		\caption{Ground Truth}
		\label{fig:sfig3}
	\end{subfigure}
	
	\begin{subfigure}{.27\linewidth}
		\centering
		\includegraphics[width=.9\linewidth]{figs/end-to-end/original_shotzero-expertA-a4775-DSC_0040_real_A}
		\caption{Input + ``color balance improved, sharpness improved"}
		\label{fig:sfig1}
	\end{subfigure}%
	\begin{subfigure}{.27\linewidth}
		\centering
		\includegraphics[width=.9\linewidth]{figs/end-to-end/original_shotzero-expertA-a4775-DSC_0040_fake_B}
		\caption{Output}
		\label{fig:sfig2}
	\end{subfigure}
	\begin{subfigure}{.27\linewidth}
		\centering
		\includegraphics[width=.9\linewidth]{figs/end-to-end/original_shotzero-expertA-a4775-DSC_0040_real_B}
		\caption{Ground Truth}
		\label{fig:sfig3}
	\end{subfigure}

	\begin{subfigure}{.27\linewidth}
		\centering
		\includegraphics[width=.9\linewidth]{figs/end-to-end/original_shotzero-expertA-a4864-jmacdscf0536_real_A}
		\caption{Input + ``saturation raised; colors are much deeper"}
		\label{fig:sfig1}
	\end{subfigure}%
	\begin{subfigure}{.27\linewidth}
		\centering
		\includegraphics[width=.9\linewidth]{figs/end-to-end/original_shotzero-expertA-a4864-jmacdscf0536_fake_B}
		\caption{Output}
		\label{fig:sfig2}
	\end{subfigure}
	\begin{subfigure}{.27\linewidth}
		\centering
		\includegraphics[width=.9\linewidth]{figs/end-to-end/original_shotzero-expertA-a4864-jmacdscf0536_real_B}
		\caption{Ground Truth}
		\label{fig:sfig3}
	\end{subfigure}
	
	\begin{subfigure}{.27\linewidth}
		\centering
		\includegraphics[width=.9\linewidth]{figs/end-to-end/original_shotzero-expertA-a4967-kme_2360_real_A}
		\caption{Input + ``color increased and skin tone of human is also good"}
		\label{fig:sfig1}
	\end{subfigure}%
	\begin{subfigure}{.27\linewidth}
		\centering
		\includegraphics[width=.9\linewidth]{figs/end-to-end/original_shotzero-expertA-a4967-kme_2360_fake_B}
		\caption{Output}
		\label{fig:sfig2}
	\end{subfigure}
	\begin{subfigure}{.27\linewidth}
		\centering
		\includegraphics[width=.9\linewidth]{figs/end-to-end/original_shotzero-expertA-a4967-kme_2360_real_B}
		\caption{Ground Truth}
		\label{fig:sfig3}
	\end{subfigure}
	
	\caption{More results for our end-to-end model.}
	\label{exp:end-to-end}
	\vspace{-0.2em} 
\end{figure*}

\section{Graph GRU Example}
\label{graphgru_ex}

 More examples with Graph RNN is given in Fig. \ref{exp:filter-bank-graphrnn}.

\begin{figure*}
	\centering
	
	\begin{subfigure}{.27\linewidth}
		\centering
		\includegraphics[width=.9\linewidth]{figs/graph_rnn_filterbank/expertA-original_shotzero-a4008-dgw_019_real_A}
		\caption{Input + ``it looked as if they lowered the sharpness of the picture , as well as the brightness and the contrast ."}
		\label{fig:sfig1}
	\end{subfigure}%
	\begin{subfigure}{.27\linewidth}
		\centering
		\includegraphics[width=.9\linewidth]{figs/graph_rnn_filterbank/expertA-original_shotzero-a4008-dgw_019_fake_B}
		\caption{Output with RNN}
		\label{fig:sfig2}
	\end{subfigure}
	\begin{subfigure}{.27\linewidth}
		\centering
		\includegraphics[width=.9\linewidth]{figs/graph_rnn_filterbank/expertA-original_shotzero-a4008-dgw_019_fake_B_graph}
		\caption{Output with Graph RNN}
		\label{fig:sfig3}
	\end{subfigure}

	\begin{subfigure}{.27\linewidth}
		\centering
		\includegraphics[width=.9\linewidth]{figs/graph_rnn_filterbank/expertA-original_shotzero-a4177-jmac_DSC1412_real_A}
		\caption{Input + ``contrast is lowered . fading is used in this edited image . they decrease the sharpness , warmth and shadows ."}
		\label{fig:sfig1}
	\end{subfigure}%
	\begin{subfigure}{.27\linewidth}
		\centering
		\includegraphics[width=.9\linewidth]{figs/graph_rnn_filterbank/expertA-original_shotzero-a4177-jmac_DSC1412_fake_B}
		\caption{Output with RNN}
		\label{fig:sfig2}
	\end{subfigure}
	\begin{subfigure}{.27\linewidth}
		\centering
		\includegraphics[width=.9\linewidth]{figs/graph_rnn_filterbank/expertA-original_shotzero-a4177-jmac_DSC1412_fake_B_graph}
		\caption{Output with Graph RNN}
		\label{fig:sfig3}
	\end{subfigure}

	\begin{subfigure}{.27\linewidth}
		\centering
		\includegraphics[width=.9\linewidth]{figs/graph_rnn_filterbank/original_shotzero-expertA-a4004-050812_113412__MG_3770_real_A}
		\caption{Input + ``the editor made the mountains pop out more ."}
		\label{fig:sfig1}
	\end{subfigure}%
	\begin{subfigure}{.27\linewidth}
		\centering
		\includegraphics[width=.9\linewidth]{figs/graph_rnn_filterbank/original_shotzero-expertA-a4004-050812_113412__MG_3770_fake_B}
		\caption{Output with RNN}
		\label{fig:sfig2}
	\end{subfigure}
	\begin{subfigure}{.27\linewidth}
		\centering
		\includegraphics[width=.9\linewidth]{figs/graph_rnn_filterbank/original_shotzero-expertA-a4004-050812_113412__MG_3770_fake_B_graph}
		\caption{Output with Graph RNN}
		\label{fig:sfig3}
	\end{subfigure}

	\begin{subfigure}{.27\linewidth}
		\centering
		\includegraphics[width=.9\linewidth]{figs/graph_rnn_filterbank/original_shotzero-expertA-a4240-_DSC0017-1_real_A}
		\caption{Input + ``looks much better than the original image . need to apply color balance to the original image"}
		\label{fig:sfig1}
	\end{subfigure}%
	\begin{subfigure}{.27\linewidth}
		\centering
		\includegraphics[width=.9\linewidth]{figs/graph_rnn_filterbank/original_shotzero-expertA-a4240-_DSC0017-1_fake_B}
		\caption{Output with RNN}
		\label{fig:sfig2}
	\end{subfigure}
	\begin{subfigure}{.27\linewidth}
		\centering
		\includegraphics[width=.9\linewidth]{figs/graph_rnn_filterbank/original_shotzero-expertA-a4240-_DSC0017-1_fake_B_graph}
		\caption{Output with Graph RNN}
		\label{fig:sfig3}
	\end{subfigure}

	\begin{subfigure}{.27\linewidth}
		\centering
		\includegraphics[width=.9\linewidth]{figs/graph_rnn_filterbank/original_shotzero-expertA-a4245-_DGW9109_real_A}
		\caption{Input + ``improve color , apply color balance , apply white balance"}
		\label{fig:sfig1}
	\end{subfigure}%
	\begin{subfigure}{.27\linewidth}
		\centering
		\includegraphics[width=.9\linewidth]{figs/graph_rnn_filterbank/original_shotzero-expertA-a4245-_DGW9109_fake_B}
		\caption{Output with RNN}
		\label{fig:sfig2}
	\end{subfigure}
	\begin{subfigure}{.27\linewidth}
		\centering
		\includegraphics[width=.9\linewidth]{figs/graph_rnn_filterbank/original_shotzero-expertA-a4245-_DGW9109_fake_B_graph}
		\caption{Ground Truth}
		\label{fig:sfig3}
	\end{subfigure}

	\caption{More results for our filter bank model with Graph RNN.}
	\label{exp:filter-bank-graphrnn}
	 
\end{figure*}